\documentclass[journal]{IEEEtran}
\usepackage{graphicx}
\usepackage{hyperref}
\usepackage{url}
\usepackage{amsmath}
\usepackage{amssymb}
\usepackage{algorithmic}
\usepackage{algorithm}
\usepackage{color}
\usepackage{setspace}

\usepackage{times}
\usepackage{epsfig}
\usepackage{multirow}
\usepackage{makecell}
\usepackage{threeparttable}
\usepackage{booktabs}
\usepackage{subfigure}
\usepackage{xspace}
\usepackage{caption}
\newcommand{\hide}[1]{}

\makeatletter
\DeclareRobustCommand\onedot{\futurelet\@let@token\@onedot}
\def\@onedot{\ifx\@let@token.\else.\null\fi\xspace}
\def\eg{{e.g}\onedot}

\def\etc{{etc}\onedot}

\def\etal{{et al}\onedot}
\makeatother

\begin{document}
\title{Joint Multi-view Face Alignment in the Wild}
\author{Jiankang~Deng,~\IEEEmembership{Student~Member,~IEEE,}
		George~Trigeorgis,
		Yuxiang~Zhou,
		and Stefanos~Zafeiriou,~\IEEEmembership{Member,~IEEE}
\thanks{This work was partially funded by EPSRC project EP/N007743/1 (FACER2VM), as well as 
by the European Community Horizon 2020 [H2020/2014-2020] under grant agreement no. 688520 (TeSLA).}
\thanks{J. Deng, S. Zafeiriou, G. Trigeorgis and Y. Zhou are with the Intelligent Behaviour Understanding Group (IBUG), the Department of Computing, Imperial College London, London SW7 2AZ, UK.}

}

\markboth{Journal of \LaTeX\ Class Files,~Vol.~4, No.~5, April~2015}%
{Shell \MakeLowercase{\textit{et al.}}: Bare Demo of IEEEtran.cls for Journals}

\maketitle

\begin{abstract}
The de facto algorithm for facial landmark estimation involves running a face detector with a subsequent deformable model fitting on the bounding box. This encompasses two basic problems: i) the detection and deformable fitting steps are performed independently, while the detector might not provide best-suited initialisation for the fitting step, ii) the face appearance varies hugely across different poses, which makes the deformable face fitting very challenging and thus distinct models have to be used (\eg, one for profile and one for frontal faces). In this work, we propose the first, to the best of our knowledge, joint multi-view convolutional network to handle large pose variations across faces in-the-wild,  and elegantly bridge face detection and facial landmark localisation tasks. Existing joint face detection and landmark localisation methods focus only on a very small set of landmarks. By contrast, our method can detect and align a large number of landmarks for semi-frontal (68 landmarks) and profile (39 landmarks) faces. We evaluate our model on a plethora of datasets including standard static image datasets such as IBUG, 300W, COFW, and the latest Menpo Benchmark for both semi-frontal and profile faces. Significant improvement over state-of-the-art methods on deformable face tracking is witnessed on 300VW benchmark. We also demonstrate state-of-the-art results for face detection on FDDB and MALF datasets. 
\end{abstract}


\IEEEpeerreviewmaketitle

\section{Introduction}
\IEEEPARstart{O}{bject} detection in computer vision has seen a huge amount of attention
in recent years~\cite{girshick2014rich,girshick2015fast,ren2015faster}. The advances in deep learning and the use of more elaborate models, such as Inception~\cite{ioffe2015batch} and ResNet~\cite{he2016deep}, have allowed for reliable
and fine-scale non-rigid object detection even in challenging scenarios.
Out of all the objects probably the most studied one is the human face.
Face detection, although having embedded in our everyday lives through the
use of digital cameras and social media, is still an extremely challenging
problem as shown by the recent survey~\cite{zafeiriou2015survey}.

Human face in images captured in unconstrained conditions (also referred to as ``in-the-wild") is a challenging object, since facial appearance can change dramatically due to extreme pose, defocus, low resolution and occlusion. Face detection ``in-the-wild" is still regarded as a challenging task. That is, considerable effort was needed in order to appropriately customise a generic object methodology, \eg Deformable Part-Based Models~\cite{mathias2014face} and Deep Convolutional Neural Networks (DCNNs)~\cite{girshick2014rich}, in order to devise pipelines that achieve very good performance in face detection~\cite{zhu2012face,mathias2014face,chen2016supervised}.

\begin{figure}[t]
\begin{center}
 \includegraphics[width=1\linewidth]{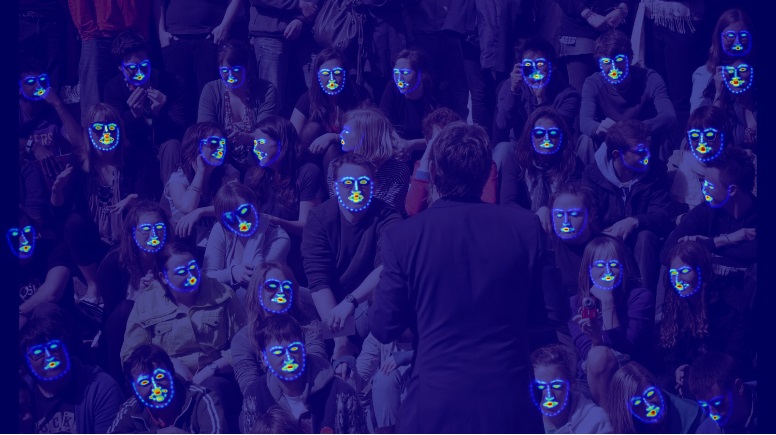}\\
\end{center}
   \caption{Facial landmark response maps generated by Multi-view Hourglass Model (MHM). The profile and frontal faces are trained jointly, and the model is robust under large pose variations.}
\label{pic:multiview_hourglass}
\end{figure}

Specifically, when dealing with human face we are also interested in detailed face alignment, that is, localising a collection of facial landmarks on face images. This step plays an important role in many face analysis task, such as face recognition~\cite{taigman2014deepface,li2015high,tai2016face}, expression recognition~\cite{bettadapura2012face,guo2016dynamic}, and face animation~\cite{thies2016face2face}. Due to the importance of the problem, a large number of facial landmark localisation methods have been proposed in the past two decades~\cite{cootes1995active,cootes2001active,cristinacce2006feature,dollar2010cascaded,cao2012face,xiong2013supervised,ren2014face,zhu2015face,sun2013deep,zhu2016face,trigeorgis2016mnemonic},
and the previous works can be categorised as parametric fitting based~\cite{cootes1995active,cootes2001active,cristinacce2006feature,tzimiropoulos2013optimization} and non-parametric regression based~\cite{dollar2010cascaded,cao2012face,xiong2013supervised,ren2014face,zhu2015face,sun2013deep,trigeorgis2016mnemonic}.
The former aims at minimising the discrepancy between the model appearance and the input image. The latter extracts
features from the image and directly regresses to the ground truth landmarks.
With the increasing number of training data~\cite{sagonas2016300}, the performance of regression-based methods is generally better than that of parametric fitting based methods.

Recently, it was shown that it is advantageous to perform jointly face detection and facial landmark localisation~\cite{chen2014joint,chen2016supervised}. Nevertheless, due to the high cost of facial landmark localisation step, only few landmarks were detected~\cite{chen2016supervised}. Furthermore, in~\cite{chen2016supervised} the method made use of extra 400K facial images from the web which are not publicly available. To avoid this, we propose a coarse-to-fine  joint multi-view landmark localisation architecture. In the coarse step, few landmarks are localised, while in the fine stage, we detect a large number of landmarks (\eg, 68/39). In our methodology, for reproducibility, we made use of publicly available data only.

\begin{figure}
\begin{center}
 \includegraphics[width=0.3\linewidth]{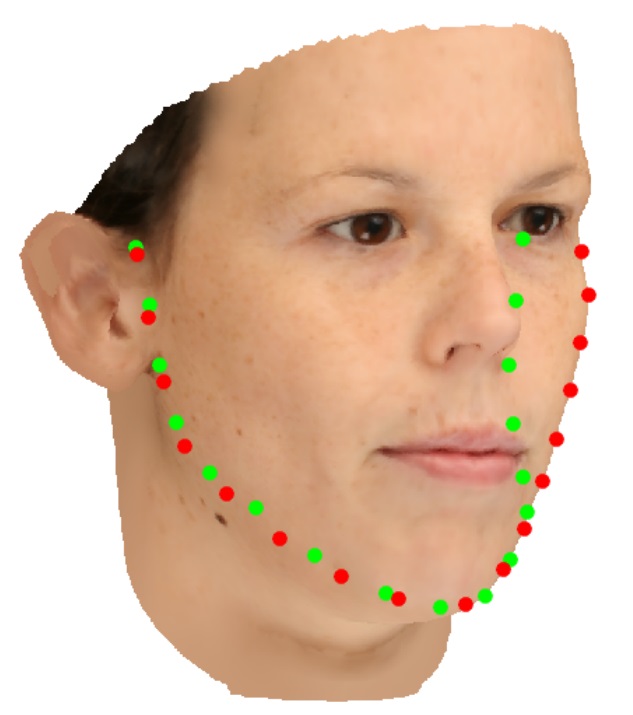}\\
\end{center}
   \caption{Inconsistent landmark annotation on face contour between 2D and 3D views. Red annotation is from 2D view, and green annotation is from 3D view.}
\label{pic:landmark_incosistent}
\end{figure}

Face alignment and tracking across medium poses, where all the landmarks are visible, has been
well addressed~\cite{xiong2013supervised,ren2014face,zhu2015face}. However, face alignment across
large poses is still a challenging problem with limited attention. There are two main challenges: Firstly, there is a controversy on landmark definition, from 2D view or 3D view? As is shown in Figure~\ref{pic:landmark_incosistent}, facial landmarks are always located at the visible face boundary in the 2D annotation. Faces which exhibit large facial poses are extremely challenging to annotate, because the landmarks on the invisible face side stack together. Since the invisible face contour needs to be always guessed to be consistent with 3D face models, labelling the self-occluded 3D landmarks is also ambiguous for annotators. Secondly, since occlusions can occur on both frontal and profile face images, designing a single shape constraint is hard for large pose face alignment. As view variation is continuous, view-specific modelling~\cite{cootes2002view,deng2016m} inevitably brings the problem of view classification and increases the computation cost.

In this work we present the first, to the best of our knowledge, method for deformable face modelling which jointly detects the face and localises a large amount of landmarks.
\begin{enumerate}
    \item We employ a coarse-to-fine strategy where a face detector is first applied to find a coarse estimate of the facial shape using a small subset of landmarks. After removing the similarity transformation, a refining step is performed to estimate the dense facial shape of each person.

    \item We formulate a novel Multi-view Hourglass Model (MHM) which tries to jointly estimate both semi-frontal and profile facial landmarks. Different from the other methods which employ distinct models, we try to capitalise on the correspondences between the profile and frontal facial shapes.

    \item We demonstrate huge improvement over the state-of-the-art results in the latest benchmarks for deformable face fitting such as IBUG, 300W, COFW and the latest Menpo Benchmark. We demonstrate state-of-the-art results for the deformable face tracking on the 300VW benchmark and face detection on FDDB and MALF.
\end{enumerate}

\section{Related work}\label{sec:related}

To better understand the problem of deformable face fitting, we review three of the major related elements.

Besides traditional models (such as AAMs~\cite{cootes2001active}, CLMs~\cite{cristinacce2006feature} and regression models \cite{xiong2013supervised,ren2016face,feng2015cascaded,liu2016dual,yang2015robust,liu2017adaptive}), recently DCNNs has been employed in face alignment \cite{sun2013deep,zhou2013extensive,liang2015unconstrained}. The resolution loss within the pooling step in DCNN was compensated by the image enlargement in a global to local way.
Zhang~\etal~\cite{zhang2014coarse} adopted the similar coarse-to-fine framework with auto-encoder networks. Ranjan~\etal~\cite{ranjan2016all} combined outputs of multi-resolution convolutional layers to predict the landmark locations.
After the presentation of the fully-convolutional network
(FCN)~\cite{liang2015unconstrained}, which takes input of arbitrary size, produces a correspondingly-sized dense label map and shows convincing results for semantic image segmentation, direct landmark coordinated prediction changed to the landmark response map prediction.
Lai~\etal~\cite{lai2016deep}, Xiao~\etal~\cite{xiao2016robust} and Bulat~\etal~\cite{bulat2016convolutional} employed the convolutional and de-convolutional network to generate the response map for each facial landmark, and added a refinement step by utilising a network that performs regression. In the area of articulated human pose estimation, Alejandro~\etal~\cite{newell2016stacked} proposed a novel stacked hourglass model, which repeated bottom-up and top-down processing in conjunction with intermediate supervision and obtained state-of-the-art result.
Bulat~\etal~\cite{bulat2017binarized} further explored binarized Hourglass-like convolutional network for face alignment with limited resources.

Despite the large volume of work on semi-frontal face alignment, literature on the large-pose scenario is rather limited. This is attributed to the fact that large-pose face alignment is a very challenging task, until now there are not enough annotated facial images in arbitrary poses (especially with a large number of landmarks). A step towards this direction is the data presented in the new facial landmark competition~\cite{stefanos2017menpo}.
The most common method in large-pose image alignment is the multi-view AAMs framework~\cite{cootes2002view}, which uses different landmark configurations for different views.
However, since each view has to be tested, the computation cost of multi-view method is always high. In \cite{zhu2012face,yu2013pose} the methods utilised the DPM framework to combine face detection and alignment, and the best view fitting was selected by the highest possibility.
Since non-frontal faces are one type of occlusions, Wu~\etal~\cite{wu2015robust} proposed a unified robust cascade regression framework that can handle both images with severe occlusion and images with large head poses by iteratively predicting the landmark visible status and the landmark locations.

To solve the problem of large pose face alignment, 3D face fitting methodologies have been considered ~\cite{jourabloo2015pose,jourabloo2016large,zhu2016face}, which
aims to fit a 3D morphable model (3DMM)~\cite{blanz2003face} to a 2D image.
\cite{jourabloo2015pose} aligned faces of arbitrary poses with the assist of a sparse 3D point distribution model. The model parameter and projection matrix are estimated by the cascaded linear or nonlinear regressors.
\cite{jourabloo2016large} extended~\cite{jourabloo2015pose} by fitting a dense 3D morphable model, employing the CNN regressor with 3D-enabled features, and estimating contour landmarks.
\cite{zhu2016face} fitted a dense 3D face model to the image via CNN and synthesised large-scale training samples in profile views to solve the problem of data labelling.
3D face alignment methods model the 3D face shape with a linear subspace and achieve fitting by minimising the difference between image and model appearance. Although 3D alignment methods can cover arbitrary poses, the accuracy of alignment is bounded by the linear parametric 3D model, and the invisible landmarks are predicted after the visible appearance are fitted. In this paper, we focus on non-parametric visible landmark localisation.

Finally, we assess our methodology for facial landmark tracking in 300VW~\cite{shen2015first}. The current state-of-the-art around face deformable tracking boils down to a pipeline which combines a generic face detection algorithm with a facial landmark localisation method~\cite{chrysos2016comprehensive}. Variants of this pipeline with different detectors or deformable models appear in the related paper~\cite{chrysos2016comprehensive}. The pipeline is quite robust since the probability of drifting is reduced due to the application of the face detector at each frame. We demonstrate that by applying the proposed methodology, large improvements over the state-of-the-art can be achieved.

\section{Our method}

\begin{figure*}[ht]
\begin{center}
 \includegraphics[width=1\linewidth]{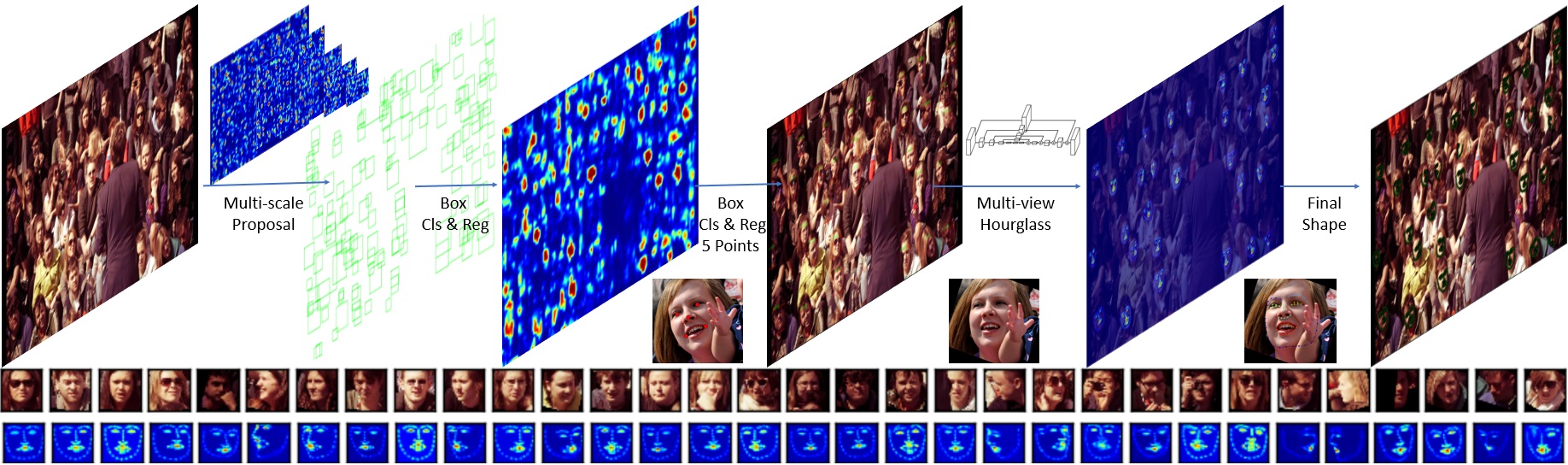}\\
\end{center}
   \caption{Proposed coarse-to-fine joint multi-view face alignment. Face regions are generated by the multi-scale proposal, then classified and regressed by the following network. Five facial landmarks are predicted to remove the similarity transformation of each face region. Multi-view Hourglass Model is trained to predict the response map for each landmark. The second and third rows show the normalised face regions and the corresponding response maps, respectively.}
\label{pic:overview}
\end{figure*}

In Figure~\ref{pic:overview}, we shown the pipeline of the proposed coarse-to-fine joint multi-view deformable face fitting method. First, face proposals are generated by a small fully convolutional network on the image pyramid. Then, these face boxes are classified and regressed to predict the five facial landmarks. Afterwards, the similarity transformation between faces are removed using the five facial landmarks, and the response map for each landmark estimate is calculated by the joint multi-view hourglass model. Lastly, we make the final prediction of each landmark based on the corresponding response map.

\subsection{Face Region Normalisation}

\begin{figure}[h]
  \centering
  \includegraphics[width=0.95\linewidth]{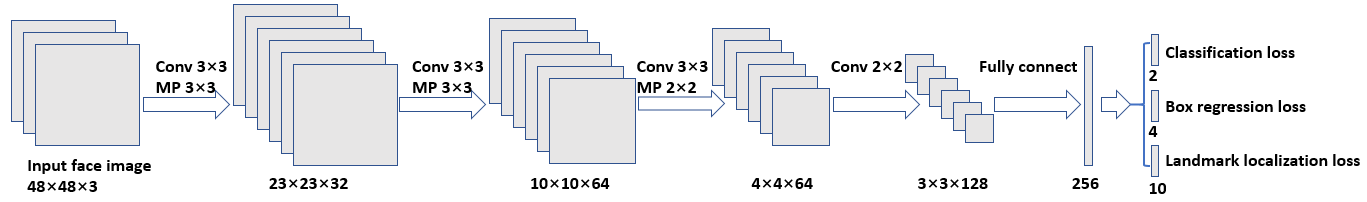}\\
  \caption{ The architecture of 3rd cascade network~\cite{zhang2016joint}. ``Conv'' means convolution, ``MP'' means max pooling, and N is the number of landmarks. The step size in convolution and pooling is 1 and 2 respectively.}\label{pic:posenet}
\end{figure}

The training of our face detection module follows the exact design of three cascade network and sampling strategies in~\cite{zhang2016joint}. In that, we minimise an objective function with the multi-task loss. For each face box $i$, its loss function is defined as:
\begin{equation}\label{eq:lossMTCNN}
L = L_{1}(p_i, p^{*}_i) + \lambda_1 p^{*}_i L_{2}(t_i, t^{*}_i) + \lambda_2 p^{*}_i L_{3} (l_i, l^{*}_i),
\end{equation}
where $p_i$ is the probability of box $i$ being a face; $p^{*}_i$ is a binary indicator (1 for positive and 0 for negative examples); the classification loss $L_{1}$ is the softmax loss of two classes (face / non-face); $t_i=\{t_x, t_y, t_w, t_h\}_i$ and $t^{*}_i=\{t^{*}_x, t^{*}_y, t^{*}_w, t^{*}_h\}_i$ represent the coordinates of the predicted box and ground truth box correspondingly. $l_i=\{l_{x_1}, l_{y_1}, \cdots , l_{x_5}, l_{y_5}\}_i$ and $l^{*}_i=\{l^{*}_{x_1}, l^{*}_{y_1}, \cdots , l^{*}_{x_5}, l^{*}_{y_5}\}_i$ represent the predicted and ground truth five facial landmarks. The box and the landmark regression targets are normalised by the face size of the ground truth.
We use $L_{2}(t_i, t^{*}_i)=R(t_i - t^{*}_i)$  and $L_{3}(l_i, l^{*}_i)=R v^{*}_i(l_i - l^{*}_i)$ for the box and landmark regression loss, respectively, where $R$ is the robust loss function (smooth-L$_1$) defined in~\cite{girshick2015fast}. In Figure~\ref{pic:posenet}, we give the network structure of the third cascade network with multi-task loss.

\begin{figure}[h]
  \centering
  \includegraphics[width=0.95\linewidth]{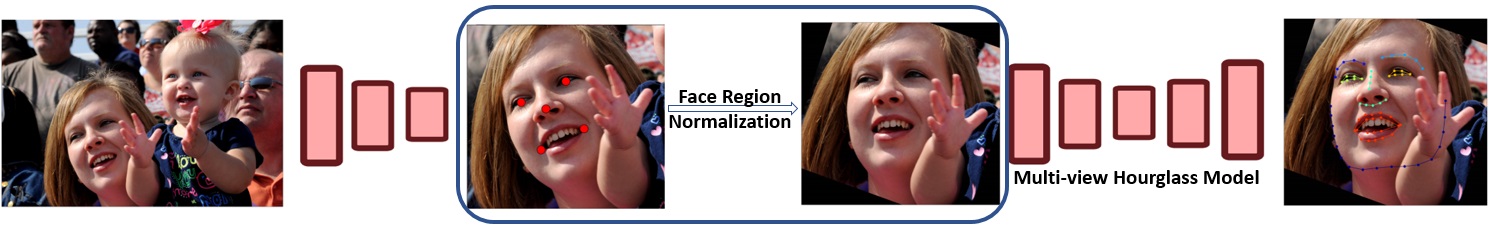}\\
  \caption{ Face Region Normalisation. The five facial landmark localisation network acts as the spatial transform layer and the subsequent alignment network is evaluated on a translation, rotation and scale invariant face image, therefore, potentially reducing the trainable parameters as well as the difficulty in learning large pose variations.}\label{pic:facenorm}
\end{figure}

One core idea of our method is to incorporate a spatial transformation~\cite{jaderberg2015spatial} which is responsible for warping the original image into a canonical representation such that the later alignment task is simplified. Recent work (\eg,~\cite{tadmor2016learning}) has explored this idea on face recognition and witnessed an improvement on the performance. In Figure~\ref{pic:facenorm}, the five facial landmark localisation network (Figure \ref{pic:posenet}) as the spatial transform layer is trained to map the original image to the parameters of a warping function (\eg, a similarity transform), such that the subsequent alignment network is evaluated on a translation, rotation and scale invariant face image, therefore, potentially reducing the trainable parameters as well as the difficulty in learning large pose variations. Since different training data are used in face region normalisation (CelebA~\cite{liu2015faceattributes} and AFLW~\cite{kostinger2011annotated}) and multi-view alignment (300W~\cite{sagonas2016300} and Menpo Benchmark~\cite{stefanos2017menpo} ), end-to-end training of these two networks with intermediate supervision on the face region normalisation step is equal to step-wise training. In this paper, we employ step-wise cascade structure, and the face region normalisation step benefits from larger training data as annotation of the five facial landmarks is much easier than dense annotation.  

\subsection{Multi-view Hourglass Model}

Hourglass~\cite{newell2016stacked} is designed based on Residual blocks~\cite{he2016deep,he2016identity}, which can be represented as follows:
\begin{equation}\label{eq:residual}
  x_{n+1} = H(x_{n})+F(x_{n},W_{n}),
\end{equation}
where $x_{n}$ and $x_{n+1}$ are the input and output of the $n$-th unit, and $F$ is the stacked convolution, batch normalisation, and ReLU non-linearity. Hourglass is a symmetric top-down and bottom-up full convolutional network. The original signals are branched out before each down-sampling step and combined together before each up-sampling step to keep the resolution information. $n$ scale Hourglass is able to extract features from the original scale to $1/2^{n}$ scale and there is no resolution loss in the whole network. The increasing depth of network design helps to increase contextual region, which incorporates global shape inference and increases robustness when local observation is blurred.

\begin{figure}[h]
\begin{center}
 \includegraphics[width=0.8\linewidth]{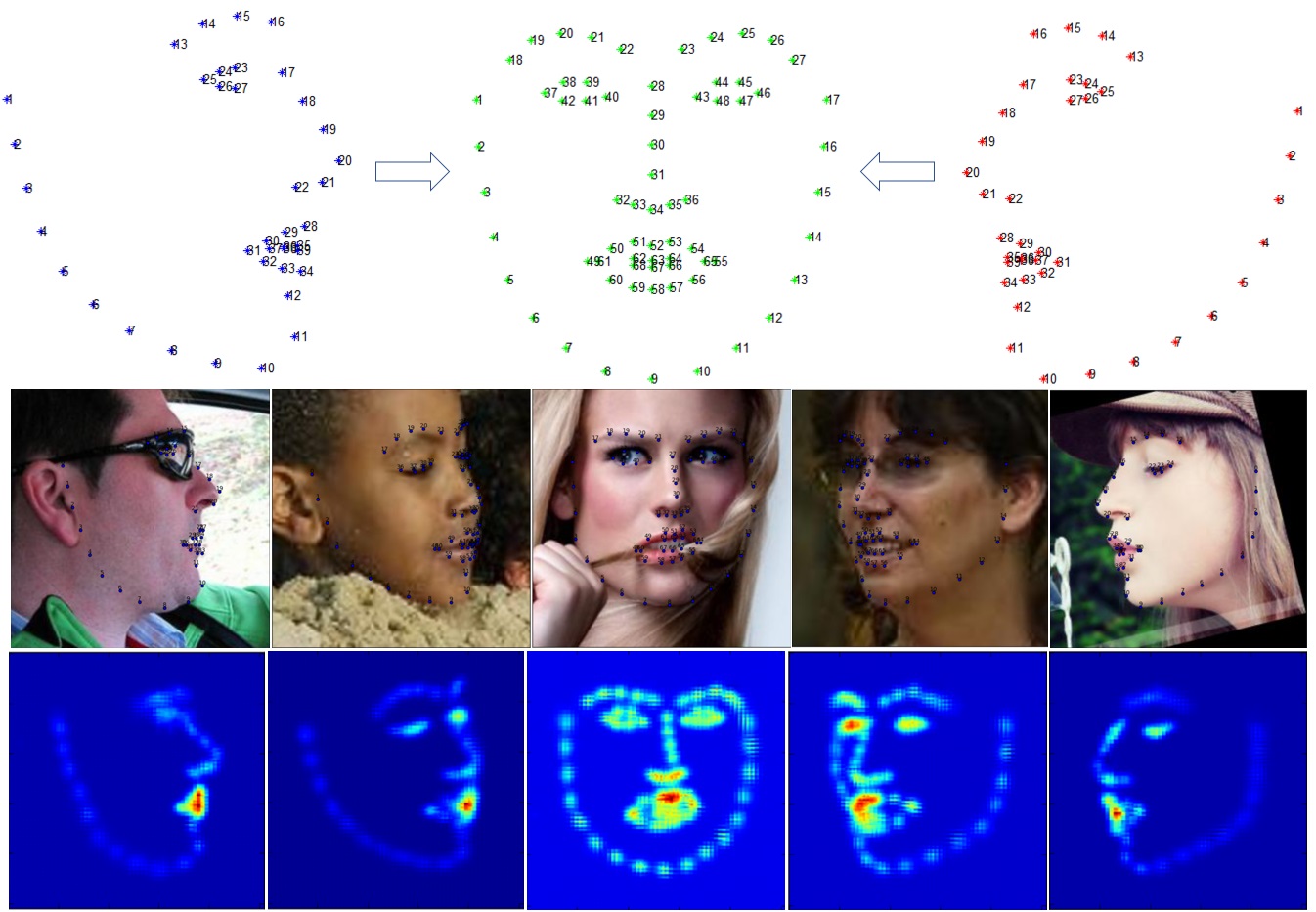}\\
\end{center}
   \caption{Multi-view Hourglass Model. First row: facial landmark configuration for frontal (68 landmarks) and profile (39 landmarks) faces~\cite{stefanos2017menpo}. We define a union landmark set with 68 landmarks for frontal and profile shape. For each landmark on the profile face, the nearest landmark on the frontal face is selected as the same definition in the union set. Third row: landmark response maps for all view faces. The response maps for semi-frontal faces (2nd and 4th) benefit from the joint multi-view training.}
\label{pic:mhm}
\end{figure}

Based on the Hourglass model~\cite{newell2016stacked}, we formulate the Multi-view Hourglass Model (MHM) which tries to jointly estimate both semi-frontal (68 landmarks) and profile (39 landmarks) face shapes. Unlike other methods which employ distinct models, we try to capitalise on the correspondences between the profile and frontal facial shapes. As shown in Figure~\ref{pic:mhm}, for each landmark on the profile face, the nearest landmark on the frontal face is regarded as its corresponding landmark in the union set, thus we can form the union landmark set with 68 landmarks (U-68). Considering that the landmark definition varies in frontal and profile data, we also enlarge the union set to 86 landmarks (U-86) by dissimilating two landmarks from eyebrow and seven landmarks from the lower part of face contour for profile annotation. During the training, we use the view status to select the corresponding response maps for the loss computation.

\begin{equation}\label{eq:lossMHM}
L = \frac{1}{N} \sum\limits_{n = 1}^N ( v_n^* \sum_{ij} {\left\| {m_n(i,j)-m_n^*(i,j)} \right\|_2^2} ),
\end{equation}
where $m_n(i,j)$ and $m_n^*(i,j)$ represent the estimated and the ground truth response maps at pixel location $(i,j)$ for the $n$-th landmark correspondingly, and $v_n\in\{0,1\}$ is the indicator to select the corresponding response map to calculate the final loss. We can see from Figure~\ref{pic:mhm} that the semi-frontal response maps (second and forth examples in third row) benefit from the joint multi-view training, and the proposed method is robust and stable in a range of poses. 

\begin{figure}[h]
  \centering
  \includegraphics[width=0.8\linewidth]{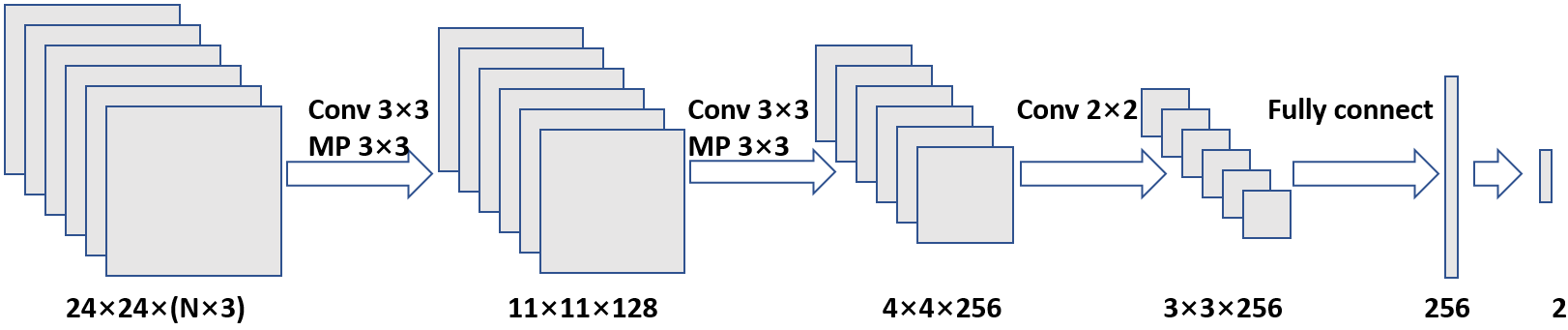}\\
  \caption{ The architecture of face classifier on the shape-indexed local patches. ``Conv'' means convolution, ``MP'' means max pooling, and N is the landmark number. The step size in convolution and pooling is 1 and 2, respectively.}\label{pic:failchecker}
\end{figure}

Based on the multi-view response maps, we extract shape-indexed patch ($24\times24$) around each predicted landmark from the down-sampled face image ($128\times128$). As shown in Figure~\ref{pic:failchecker}, a small classification network is trained to classify face / non-face. This classifier is not only used to remove high score false positives for face detection, but also can be employed as a failure checker for deformable face tracking.

\section{Experiments}

\subsection{Experiment Setting}

\subsubsection{Training Data}

\noindent \textbf{Face Detection Model:} The face detection module before the multi-view face alignment step follows the cascaded network design and sampling strategies as in~\cite{zhang2016joint}. We crop positive faces (IoU $ > 0.6$), negative faces (IoU $ < 0.3$) and part faces (IoU $ \in (0.4,0.65)$) from Wider Face~\cite{yang2016wider} training set. To guarantee a high accuracy in predicting five facial landmarks, we employ additional labelled faces from the AFLW~\cite{kostinger2011annotated} dataset besides labelled faces from CelebA~\cite{liu2015faceattributes}. For the additional classifier after the multi-view alignment step, the positive (IoU $>0.5$) and negative samples (IoU $<0.3$) are generated from the previous cascaded face detector.
\noindent \textbf{Multi-view Hourglass Model:} We train the face alignment module MHM on the 300W database~\cite{sagonas2016300}, and the Menpo Benchmark database~\cite{stefanos2017menpo}, where faces are manually annotated with either 68 (semi-frontal face) or 39 (profile face) landmarks. The training set of the 300W database (we denote as {\em 300W-68}) consists of the LFPW trainset~\cite{belhumeur2013localizing}, the Helen trainset~\cite{le2012interactive} and the AFW dataset~\cite{zhu2012face}, hence, a total of 3148 images are available. The Menpo Benchmark database~\cite{stefanos2017menpo} (denoted as {\em Menpo-39-68}) consists of 5658 semi-frontal face images and 1906 profile face images. In this paper, we defined two training sets ({\em 300W-68-Menpo-39} and {\em 300W-68-Menpo-39-68}) for different evaluation purposes. {\em 300W-68-Menpo-39} includes the {\em 300W-68} data and the profile faces of {\em Menpo-39}, while {\em 300W-68-Menpo-39-68} groups all the available training images in {\em 300W-68} and {\em Menpo-39-68}.

\subsubsection{Testing data}

\noindent \textbf{Face detection:} We evaluate the performance of our face detection module in two challenging datasets, FDDB and MALF. FDDB consists of 5171 faces in 2845 images from the unconstrained environment. MALF is a fine-grained evaluation dataset, in total, there are 5250 images with 11931 annotated faces. The ``hard'' subset contains faces (larger than $60\times60$) with huge variations in pose, expression, or occlusion. In particular, we give detailed pose-specific evaluations on MALF.
\noindent \textbf{Face alignment in images \& videos:} Evaluations of single face alignment and face tracking are performed in several {\em in-the-wild} databases. For alignment in static image, we test on {\em IBUG} dataset, {\em 300W} testset~\cite{sagonas2016300}, {\em COFW~\cite{burgos2013robust,ghiasi2015occlusion}}, and {\em Menpo-test~\cite{stefanos2017menpo}}. All these databases are collected under fully unconstrained conditions and exhibit large variations in pose, expression, illumination, ~\etc. In particular, {\em Menpo-test~\cite{stefanos2017menpo}} collects faces of all different poses, which are categorised into 5535 semi-frontal faces and 1946 profile faces based on~\cite{stefanos2017menpo}. For face tracking experiment, {\em 300VW} is the only publicly available {\em in-the-wild} benchmark. It consists of 114 videos (about 218k frames in total), captured in the wild with large pose variations, severe occlusions and extreme illuminations.

\subsubsection{Evaluation Metric}

Given the ground truth, the landmark localisation performance can be evaluated by Normalised Mean Error (NME), and the normalisation is typically carried out with respect to face size.
\begin{equation}
err = \frac{1}{M}\sum\limits_{i = 1}^M {\frac{{\frac{1}{N}\sum\limits_{j = 1}^N {{{\left| {{p_{i,j}} - {g_{i,j}}} \right|}^2_2}} }}{{{d_{i}}}}},
\end{equation}
where {\em M} is the number of images in the test set, {\em N }is the number of landmarks, {\em p} is the prediction, {\em g} is the ground truth, and {\em d} is the normalise distance. According to the protocol of difference facial alignment benchmarks, various normalise distances are used in this paper, such as eye centre distance~\cite{ren2014face}, outer eye corner distance~\cite{sagonas2016300} and diagonal distance of ground truth bounding box~\cite{chrysos2016comprehensive}. The permissible error (localisation threshold) is taken as a percentage of the normalise distance. 

\subsubsection{Training of Multi-view Hourglass Model}

The training of the proposed method follows a similar design as in the Hourglass Model~\cite{newell2016stacked}. Before the training, several pre-processing steps are undertaken. We firstly remove scale, rotation and translation differences by five facial landmarks among the training face images (referred as the spatial transformer step), then crop and resize the face regions to $256\times256$. We augment the data with rotation (+/- 30 degrees), scaling (0.75-1.25), and translation (+/- 20 pixels) that would help simulate the variations from face detector and five landmark localisation. The full network starts with a $7\times7$ convolutional layer with stride 2, followed by a residual module and a round of max pooling to bring the resolution down from 256 to 64, as it could save GPU memory while preserving alignment accuracy. The network is trained using Tensorflow~\cite{abadi2016tensorflow} with an initial learning rate of 1e-4, batch size of 12, and learning steps of 100k. The Mean Squared Error (MSE) loss is applied to compare the predicted heatmaps to the ground-truth heatmaps. Each training step takes 1.2s on one NVIDIA GTX Titan X (Pascal) GPU card. During testing, face regions are cropped and resized to $256\times256$, and it takes 12.21ms to generate the response maps. 

\subsection{Ablation Study}\label{subsec:ablation}

We consider different training strategies and validate these setting on the challenging IBUG dataset in Table~\ref{table:IBUG}.
(1) Hourglass Model (HM) trained on {\em 300W-68}.
(2) HM trained on {\em 300W-68}, with spatial transformer step based on five facial landmarks.
(3) HM trained on {\em 300W-68} with simulated response maps from the output five landmarks. The input channel increases from 3 to 8, and this Hourglass model is trained with the spatial facial clue from face detector. The result of Method 3 is worse than that of Method 2, which indicates that the spatial transformer step for each face region is better than the spatial indication.
(4) Multi-view Hourglass Model (MHM) trained on {\em 300W-68-Menpo-39} with 68 union landmarks.
(5) MHM trained on {\em 300W-68-Menpo-39} with 86 union landmarks.
(6) MHM trained on {\em Menpo-39-68} with 68 union landmarks.
(7) MHM trained on {\em 300W-68-Menpo-39-68} with 68 union landmarks.
(8) Two-stage Multi-view Hourglass with intermediate supervision. This model barely improves the performance but doubling the computation cost.

\begin{table}[h!]
\begin{center}
\begin{tabular}{c|c|c|c}
\hline
Method & AUC & FR ($\%$)&NME ($\%$)   \\
\hline\hline
1 &0.4470 &8.14 &6.09   \\
2 &0.4737 &1.48 &5.30   \\
3 &0.4629 &2.96 &5.49   \\
4 &0.5076 &0.74 &4.92   \\
5 &0.5141 &0.74 &4.86   \\
6 &0.5226 &0.74 &4.78   \\
7 &0.5324 &0.74 &4.68   \\
8 &0.5409 &0.74 &4.59   \\
\hline
\end{tabular}
\end{center}
\caption{Landmark localisation results on the IBUG dataset using 68 landmarks. Accuracy is reported as the Area
Under the Curve (AUC) of the Cumulative Error Distribution curve, the Failure Rate (FR) at threshold $0.1$, and out eye corner distance Normalised Mean Error (NME).}\label{table:IBUG}
\end{table}

\begin{figure}[h!]
  \centering
  \includegraphics[width=0.49\textwidth]{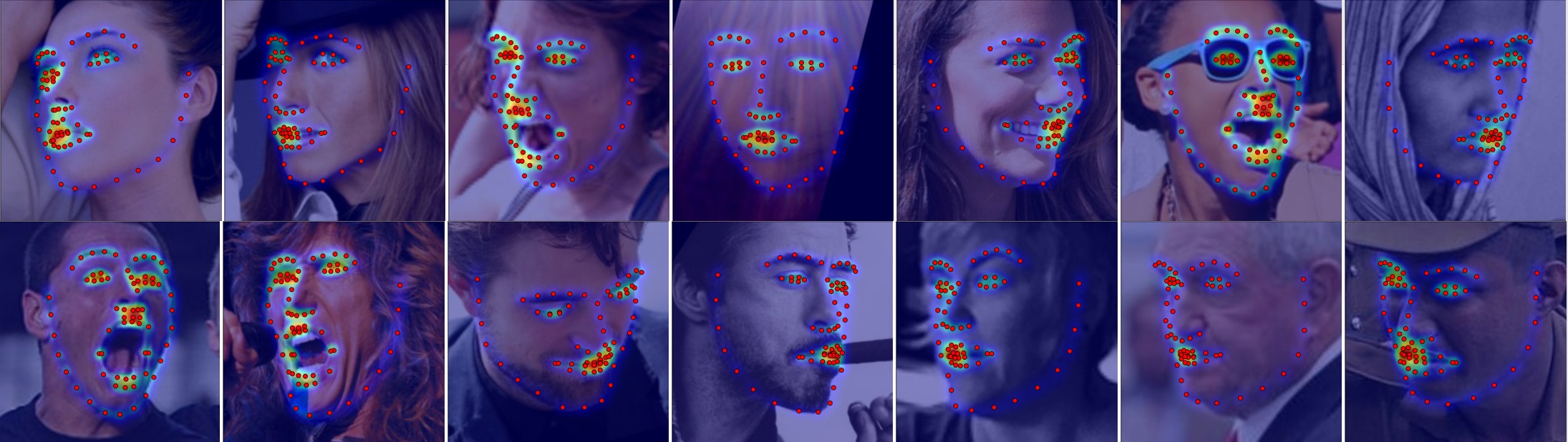}\\
  \caption{Demo results with large pose variation on IBUG predicted by Method (7). The score is higher on the inner facial organs than on the face contour.}\label{pic:IBUG_demo}
\end{figure}

From the ablation experiments, we could conclude that by integrating the spatial transformer step, joint multi-view training and feeding more quality training data, the robustness and accuracy of proposed method improve hugely. As shown in Figure~\ref{pic:IBUG_demo}, although responses are more evident on facial organs than those on face contour, owing to more available profile training data, the proposed joint Multi-view Hourglass Model is able to deal with large pose variation.

\subsection{Face Alignment on Images}

\begin{figure}[h!]
 \centering
 \includegraphics[width=0.49\textwidth]{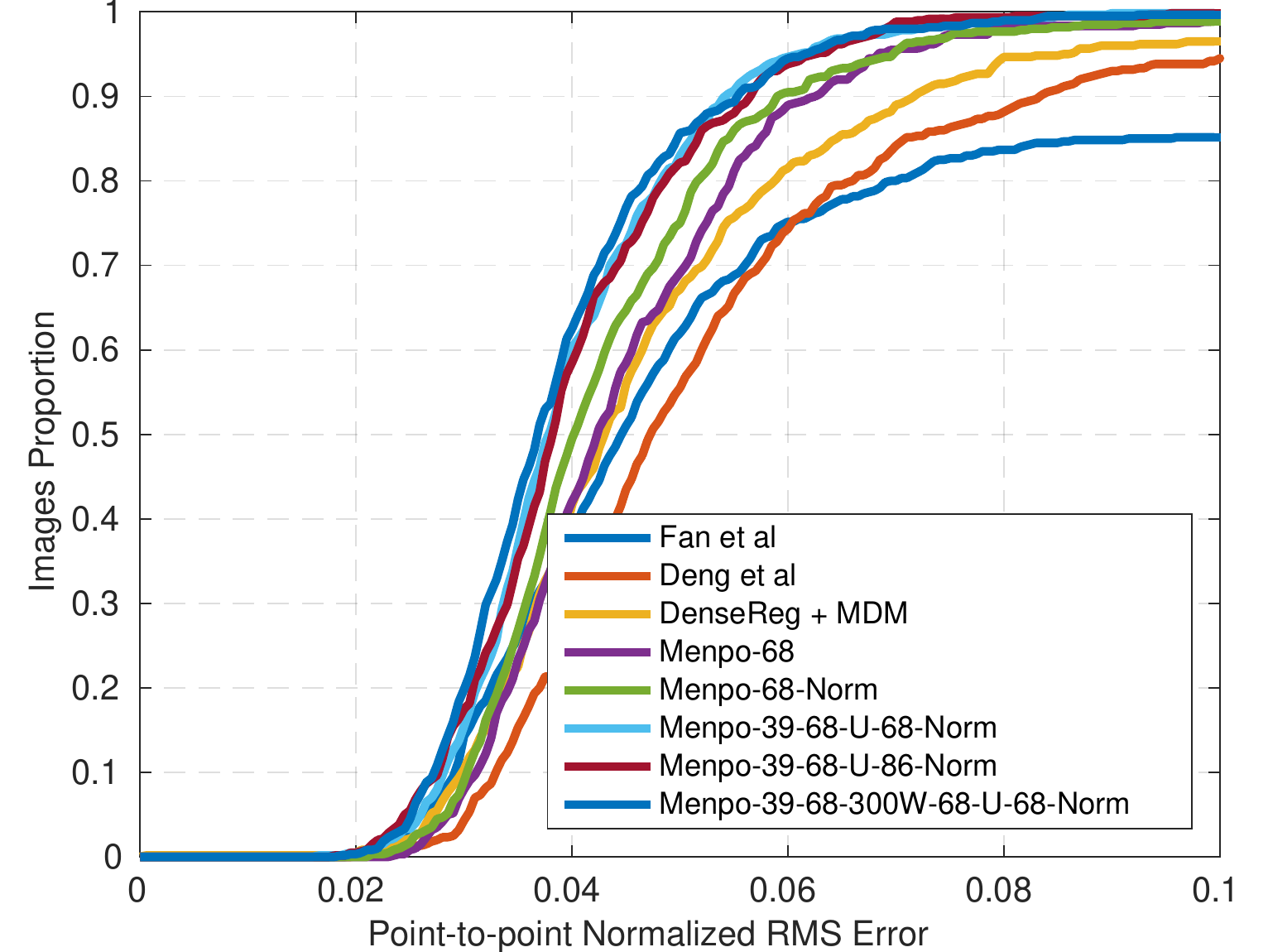}\\
 \caption{Landmark localisation results on the 300W dataset. Accuracy is reported as Cumulative Error Distribution of RMS point-to-point error normalised with the out eye corner distance.}\label{fig:300WCED}
\end{figure}

We present experimental results on three face image databases, 300W database~\cite{sagonas2016300}, COFW~\cite{burgos2013robust,ghiasi2015occlusion} dataset and Menpo Benchmark~\cite{stefanos2017menpo}. The alignment method we evaluate here is the proposed Multi-view Hourglass Model (MHM), where the \textbf{-Norm} means the spatial transformer, and the \textbf{-U-86} means the union 86 landmarks. Experiment results on 300W database are shown in Figure~\ref{fig:300WCED}, where we compared the proposed methods with the best results in the 300W competition~\cite{sagonas2016300}, such as Deng~\etal~\cite{deng2016m} and Fan~\etal~\cite{fan2016approaching}. Besides, we also compare with the state-of-the-art face alignment method ``DenseReg + MDM''~\cite{guler2016densereg}. It is obvious that our model (Menpo-39-68-300W-68-U-68-Norm) outperforms those methods by a large margin. Table~\ref{table:300W} reports the area under the curve (AUC) of the CED curves, as well as the failure rate for a maximum error of $0.1$. Apart from the accuracy improvement shown by the AUC, we believe that the reported failure rate of $0.33\%$ is remarkable and highlights the robustness of our MHM. Additionally, we found that the union landmark definition only has little influence on semi-frontal face alignment accuracy. Thus we stick to the union 68 landmarks definition to avoid any confusion.

\begin{table}[h!]
\begin{center}
\begin{tabular}{c|c|c}
\hline
Method & AUC & FR ($\%$)  \\
\hline\hline
Fan \etal   & 0.4802&14.83 \\
Deng \etal & 0.4752&5.5\\
\hline\hline
DesenReg + MDM& 0.5219& 3.67  \\
\hline\hline
Menpo-68& 0.5485&1.00\\
Menpo-68-Norm&0.5656&1.17\\
Menpo-39-68-U-68-Norm& 0.5973& 0.17  \\
Menpo-39-68-U-86-Norm& 0.5987&0.17\\
Menpo-39-68-300W-U-68-Norm& 0.6071& 0.33  \\
\hline
\end{tabular}
\end{center}
\caption{Landmark localisation results on the 300W (indoor and outdoor) testing dataset using 68 landmarks. Accuracy is reported as the Area
Under the Curve (AUC) and the Failure Rate of the Cumulative Error Distribution of the RMS point-to-point error normalised with out eye corner distance. ``Norm'' stands for the spatial transformer step from five facial landmarks. ``U'' stands for the union set number of profile and frontal data.}\label{table:300W}
\end{table}

\begin{figure}[ht!]
 \centering
 \includegraphics[width=0.49\textwidth]{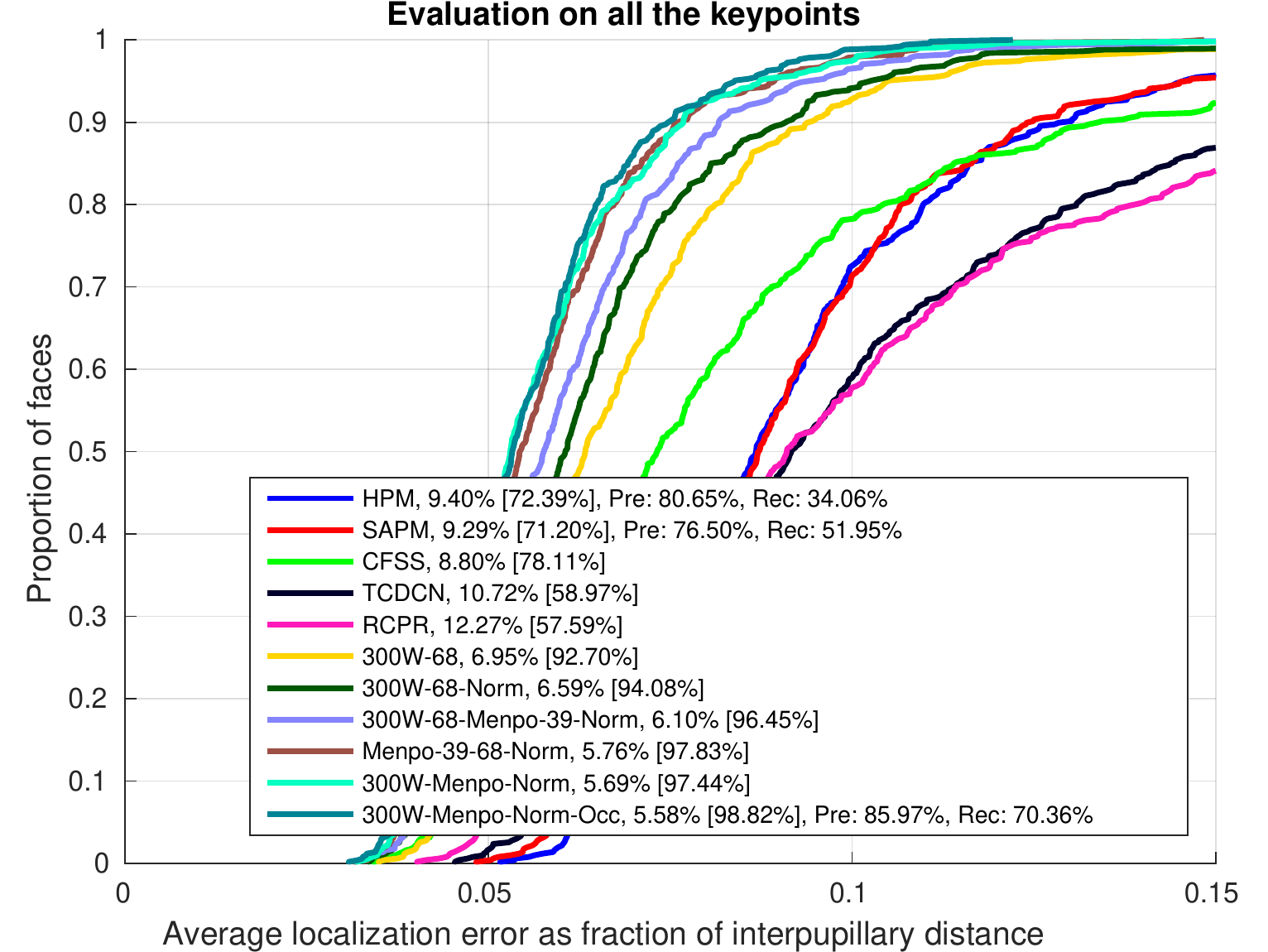}\\
 \caption{Landmark localisation results on the COFW dataset. Accuracy is reported as Cumulative Error Distribution of RMS point-to-point error normalised with the eye centre distance.} \label{fig:COFW}
\end{figure}

\begin{figure}[h!]
  \centering
  \includegraphics[width=0.49\textwidth]{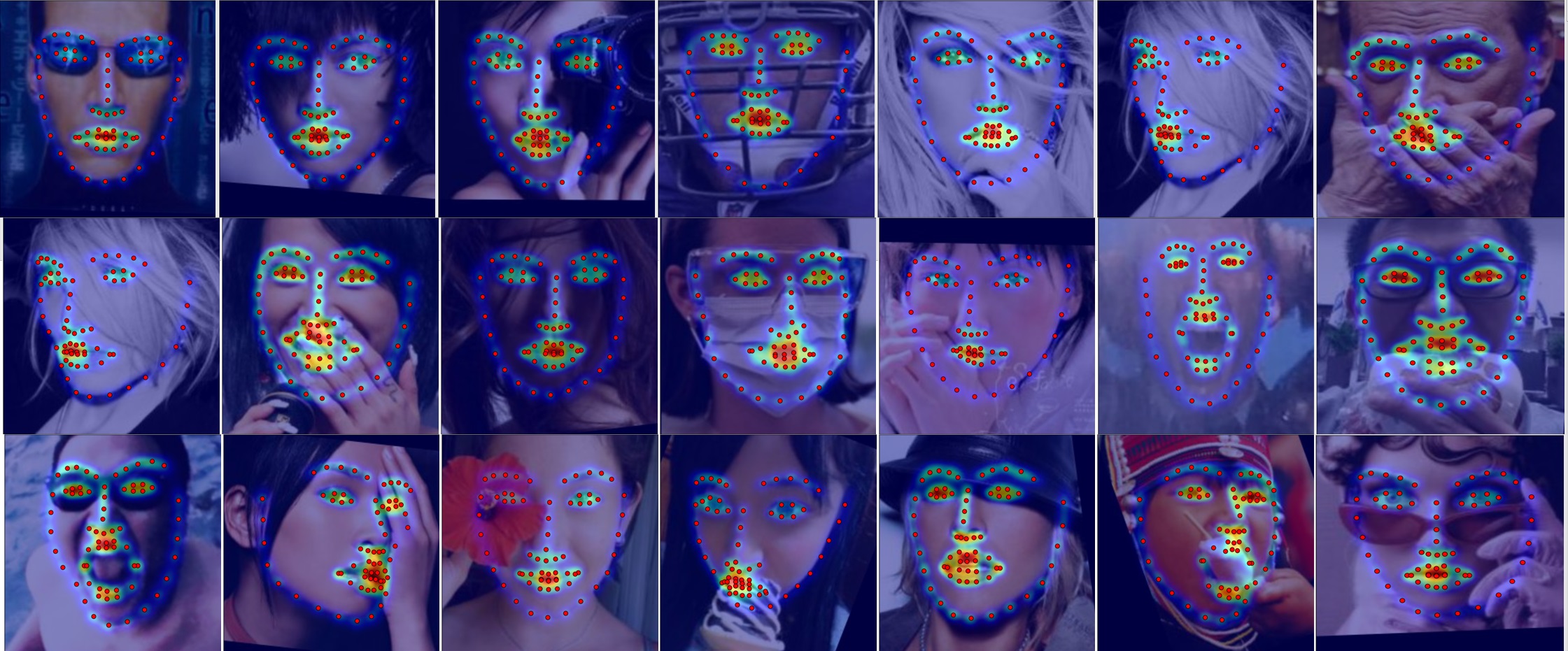}\\
  \caption{Example results by MHM on COFW. Response maps on the occluded parts are still very clear and evident.}\label{pic:COFW_demo}
\end{figure}

We also present the performance of the MHM on the COFW~\cite{burgos2013robust,ghiasi2015occlusion} dataset. Robust face alignment under occlusion and occluded landmark prediction are coupled problem that could be resolved simultaneously. Given the landmark occlusion status, local observation noise can be removed and the occluded landmark location can be predicted by shape context or constraint.
Given a good fitting result, exploiting the fact that appearance of occluded region is quite different from the normal face appearance, even the simplest binary classifier could achieve excellent performance on occlusion classification. In Figure~\ref{fig:COFW}, we show the result of the proposed method comparing with state-of-the-art methods on COFW~\cite{burgos2013robust}, such as HPM~\cite{ghiasi2014occlusion}, SAPM~\cite{ghiasi2015using}, CFSS~\cite{zhu2015face}, TCDCN~\cite{zhang2016learning}, and RCPR~\cite{burgos2013robust}. It can be clearly seen that even the baseline Hourglass model obtains a much better result because the bottom-up and top-down processing steps model the scale variations that would benefit the context inference. Moreover, by adding the spatial transformer, joint multi-view training and combined training data step-by-step, we gradually improved the alignment result, with the final success rate approaching $97.44\%$. Based on our best result, we employ the adaptive exemplar dictionary method~\cite{liu2016adaptive} to predict occlusion status and refine the occluded landmarks. The normalised mean error decreases from $5.69\%$ to $5.58\%$, and the occlusion prediction obtains a recall rate of $70.36\%$ at the precision rate of $85.97\%$. In Figure~\ref{pic:COFW_demo}, we give some fitting examples on COFW under heavy occlusions. To our surprise, responses of the occluded parts are still very clear and evident, which would prevent weird fitting results. This suggests that the proposed method captures and consolidates information across whole face images under different conditions, and incorporates local observation and global shape context in an implicit data-driven way, and thus improves the model's robustness under occlusions.

\begin{figure}[ht!]
  \centering
  \subfigure[Menpo Semi-frontal]{
    \label{fig:MenpoCED_frontal} 
    \includegraphics[width=0.49\textwidth]{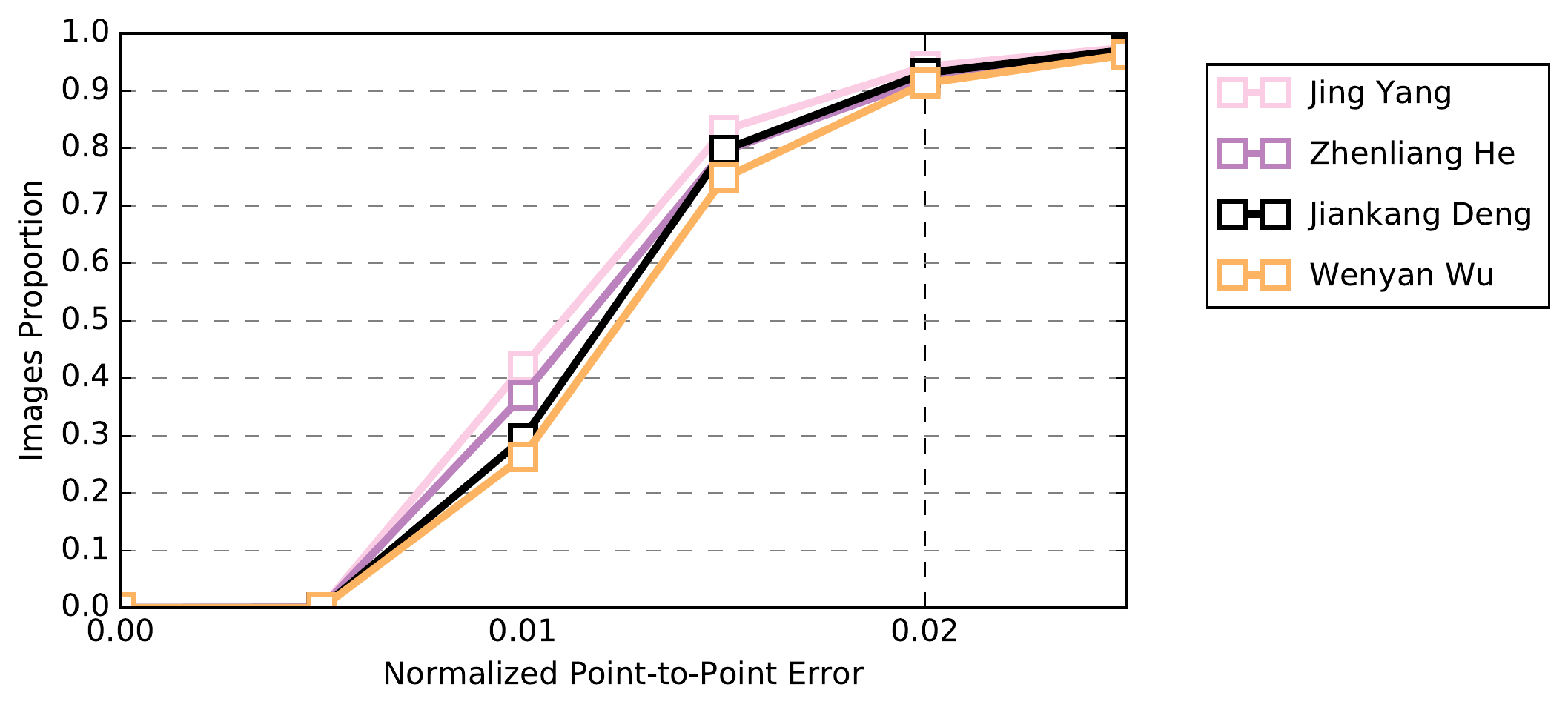}}
  \subfigure[Menpo profile]{
    \label{fig:MenpoCED_profile} 
    \includegraphics[width=0.49\textwidth]{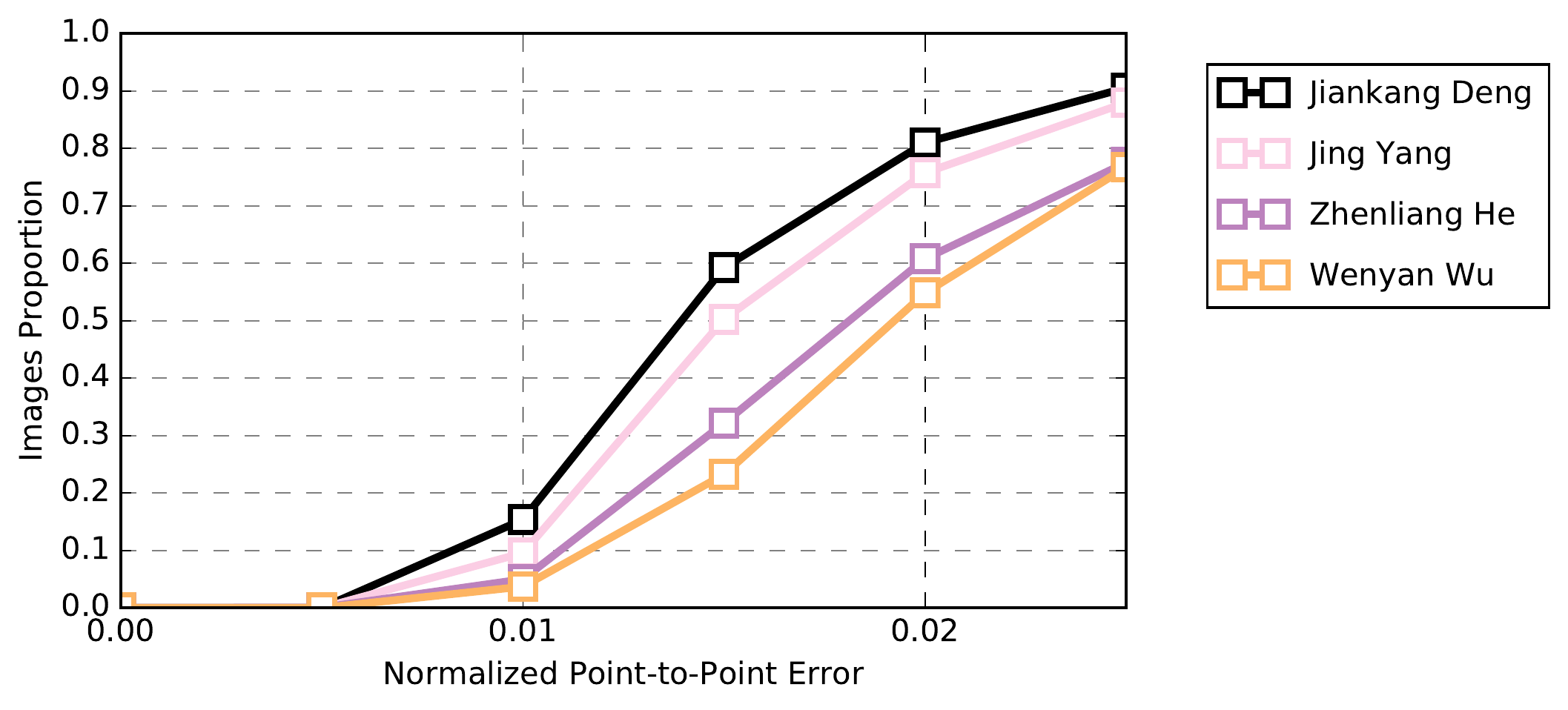}}
 \caption{Landmark localisation results on the Menpo Benchmark. Accuracy is reported as Cumulative Error Distribution of RMS point-to-point error normalised with the diagonal of the ground truth bounding box.}
\label{fig:menpoCED}
\end{figure}

\begin{figure*}
  \centering
  \subfigure[Menpo Semi-frontal set]{
    \label{fig:Menpodemo_frontal} 
    \includegraphics[width=1\linewidth]{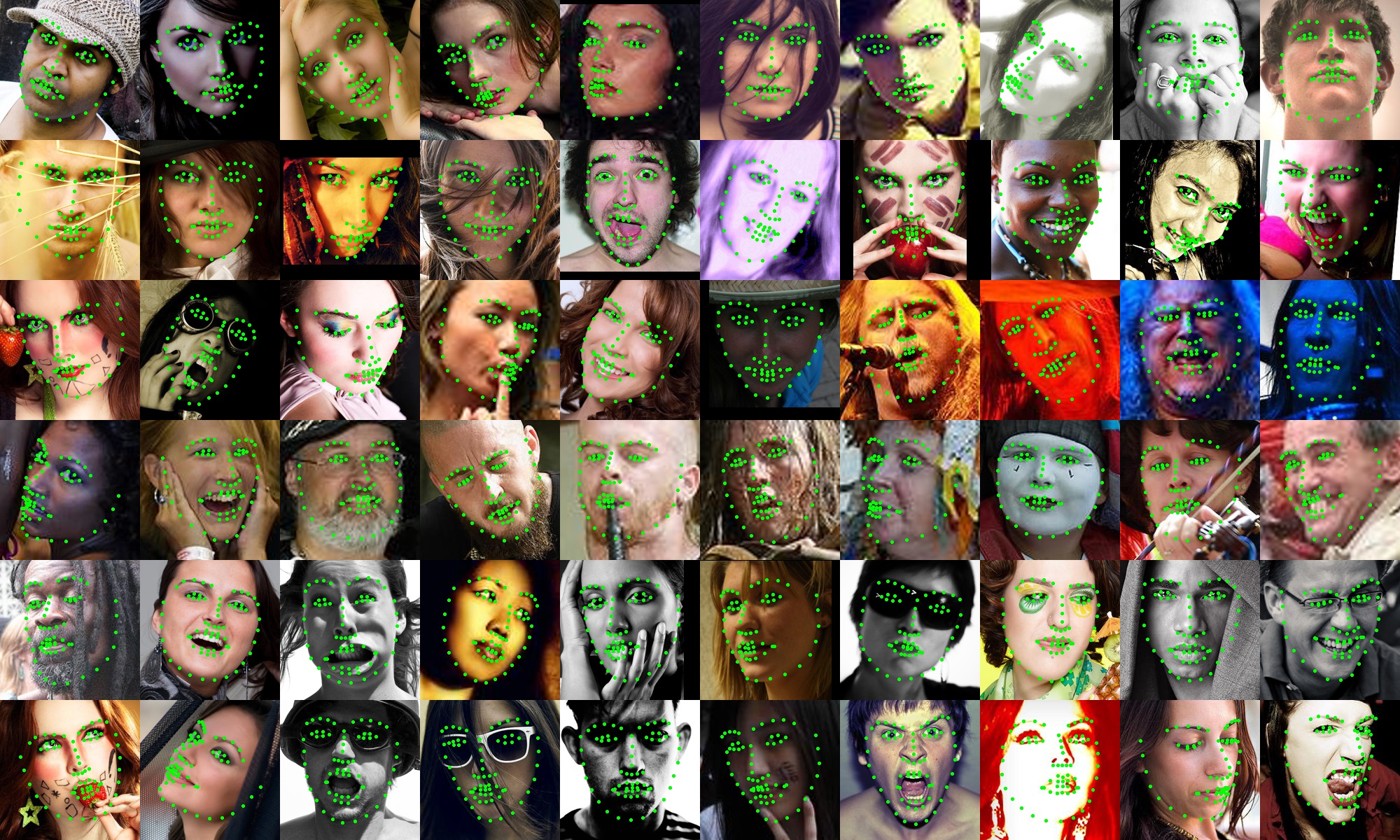}}
  \subfigure[Menpo profile set]{
    \label{fig:Menpodemo_profile} 
    \includegraphics[width=1\linewidth]{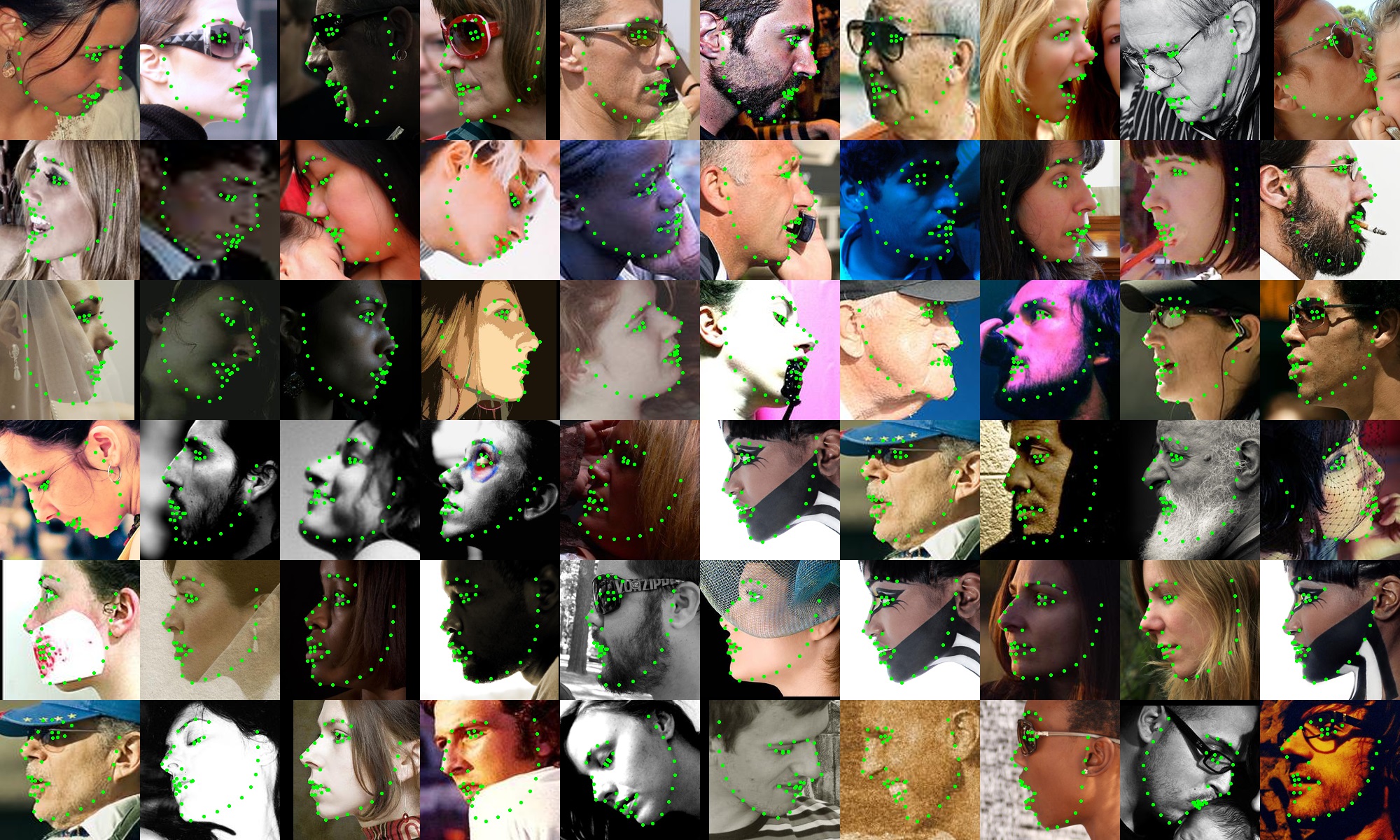}}
 \caption{Example landmark localisation results on the test set of the Menpo Benchmark.}
\label{pic:menpoDemoImage}
\end{figure*}

In Figure~\ref{fig:menpoCED}, we also report the test results of our model on the Menpo Benchmark by comparing with the best three entries (Jing Yang~\cite{yang2017stacked}, Zhenliang He~\cite{he2017robust}, Wenyan Wu~\cite{wu2017leveraging}) of the competition~\cite{stefanos2017menpo}. We draw the curve of cumulative error distribution on semi-frontal and profile test data separately. The proposed method has similar performance to the best performing methods in semi-frontal faces. Nevertheless, it outperforms the best performing method in profile faces. Despite that result on profile data is worse than that on semi-frontal data, both of their normalised (by diagonal length of bounding box) fitting errors of our method are remarkably small, approaching $1.48\%$ and $1.27\%$ for profile and semi-frontal faces respectively. In Figure~\ref{pic:menpoDemoImage}, we give some fitting examples on the Menpo test set. As we can see from the alignment results, the proposed multi-view hourglass model is robust under pose variations, exaggerate expressions and occlusions on both semi-frontal and profile subset.

\subsection{Face Alignment on Videos}

\begin{figure*}
  \centering
  \subfigure[Scenario1]{
    \label{fig:subfig:Scenario1} 
    \includegraphics[width=0.32\textwidth]{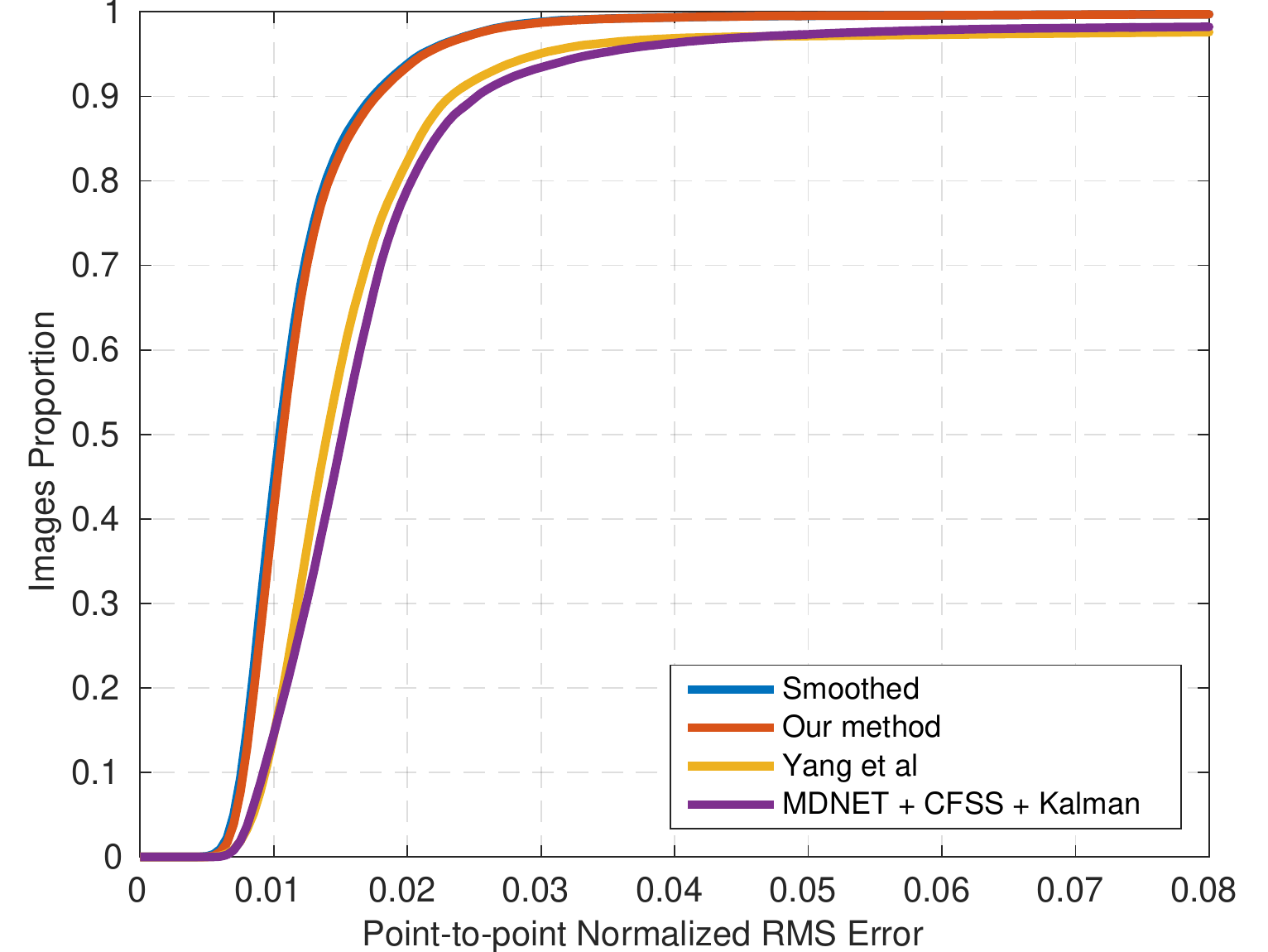}}
 \subfigure[Scenario2]{
    \label{fig:subfig:Scenario2} 
    \includegraphics[width=0.32\textwidth]{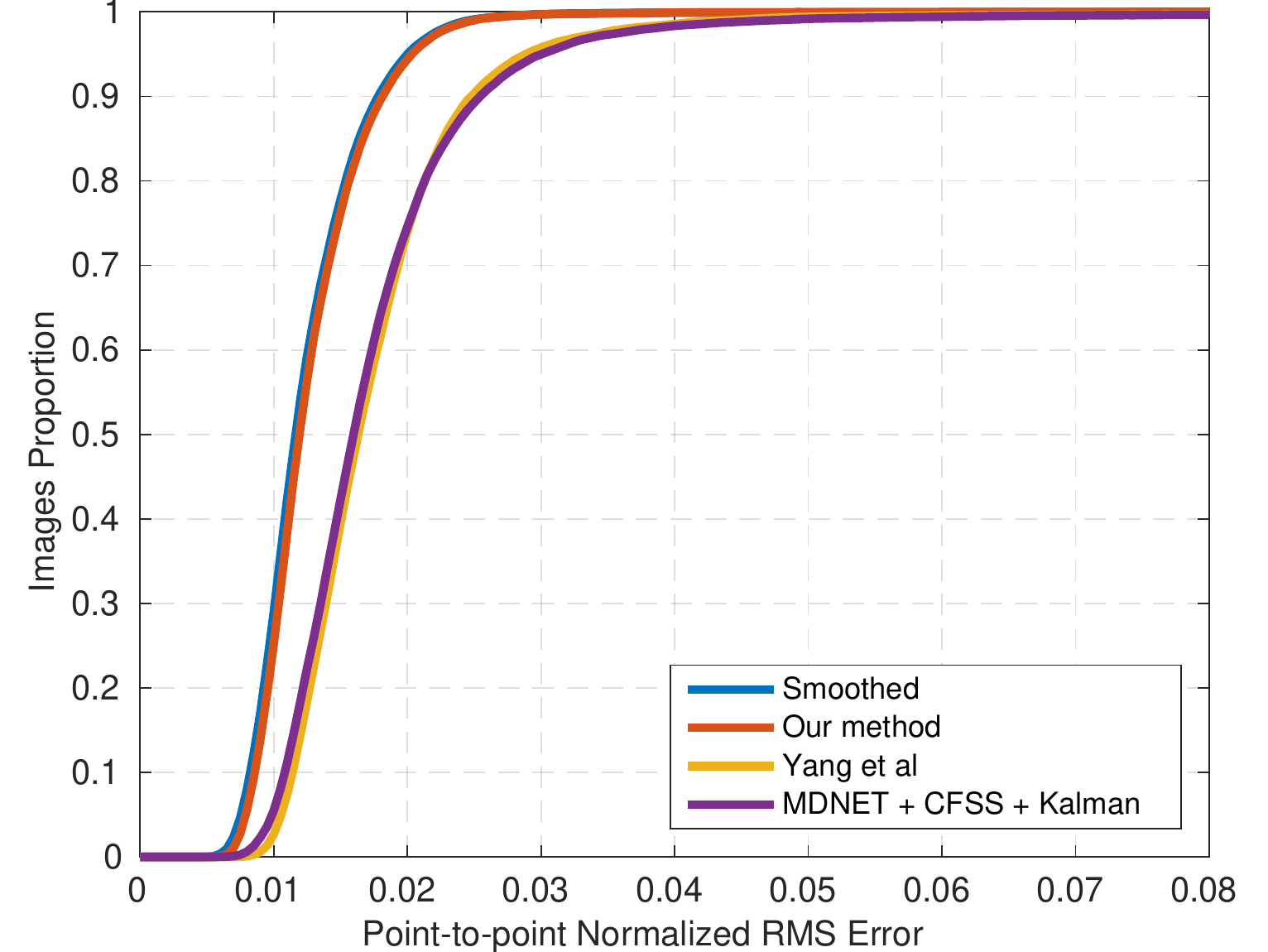}}
  \subfigure[Scenario3]{
    \label{fig:subfig:Scenario3} 
    \includegraphics[width=0.32\textwidth]{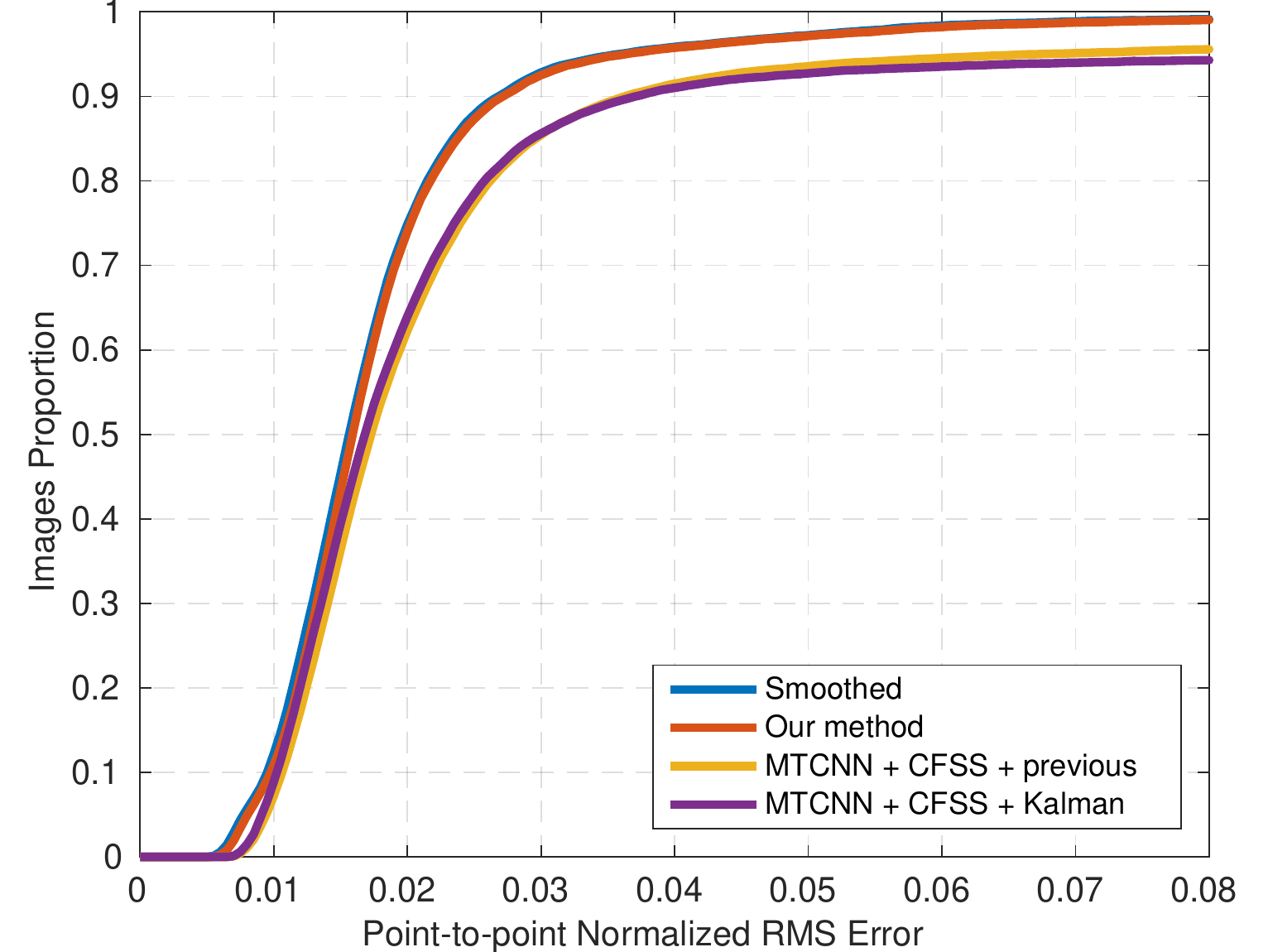}}
 \caption{Deformable face tracking results on 300VW. We only compare with the best two results evaluated by~\cite{chrysos2016comprehensive} on each scenario.}
\label{fig:300VW}
\end{figure*}

We employ the 300VW challenge~\cite{shen2015first} testset for the challenging task of deformable face tracking on videos. Using our joint MHM method, We perform a frame-by-frame tracking on the video, and we initialise the next frame by the previous facial bounding box. The classifier based on the multi-view response maps is used as the failure checker during tracking. The face detector will be called if the fitting fails. The MHM takes 12.21 ms per face, and the classifier takes 2.32ms per face. The proposed multi-view face alignment and tracking method can run at about 50 FPS on the 300VW testset. We compare our method against the winners of the 300VW challenge: Yang~\etal~\cite{yang2015facial} and Xiao~\etal~\cite{xiao2015facial}. Figure~\ref{fig:300VW} reports the CED curves for all three video scenarios, and Table~\ref{table:300VW} reports the AUC and Failure Rate measures. The proposed MHM achieves the best performance, by a large margin compared to the winner of the 300VW competition ($\ge$15\% at RMSE = 0.02 in Scenario1\&2, $\approx$10\% at RMSE = 0.02 in Scenario3) as well as the best setting for CFSS method~\cite{zhu2015face,chrysos2016comprehensive} ($\approx$15\% at RMSE = 0.02 in Scenario1\&2, $\approx$10\% at RMSE = 0.02 in Scenario3), despite the fact that our approach is not fine-tuned on the training set of 300VW, while the rest of the methods were trained on video sequences and sometimes even with temporal modelling. Besides, our frame-by-frame tracking result is good enough that additional smoothing step (Kalman Filter) might be unnecessary.

\begin{figure}[h!]
  \centering
  \subfigure[Scenario3-411]{
    \label{fig:subfig:Scenario3411} 
    \includegraphics[width=0.49\textwidth]{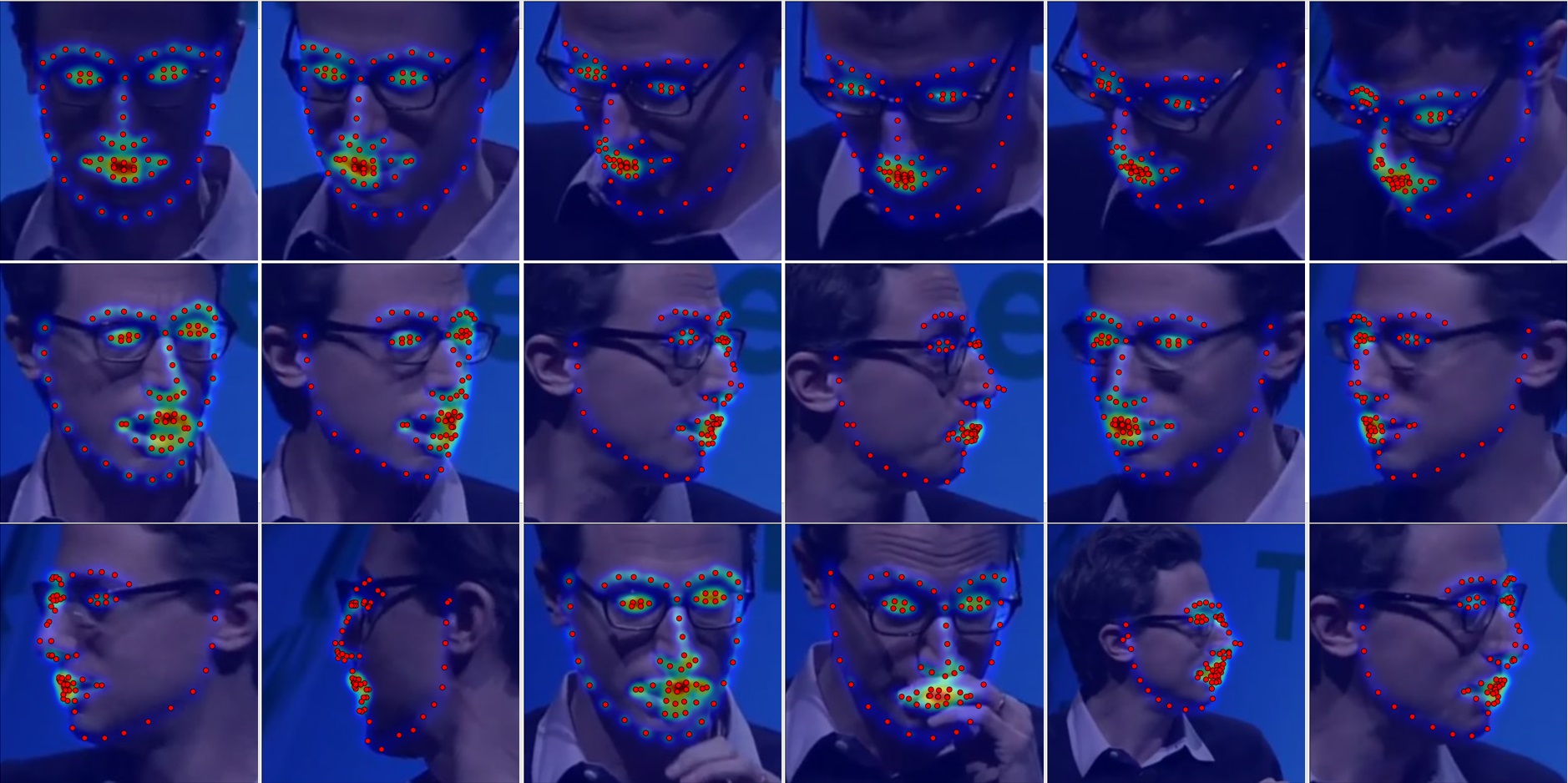}}
 \subfigure[Scenario3-557]{
    \label{fig:subfig:Scenario3557} 
    \includegraphics[width=0.49\textwidth]{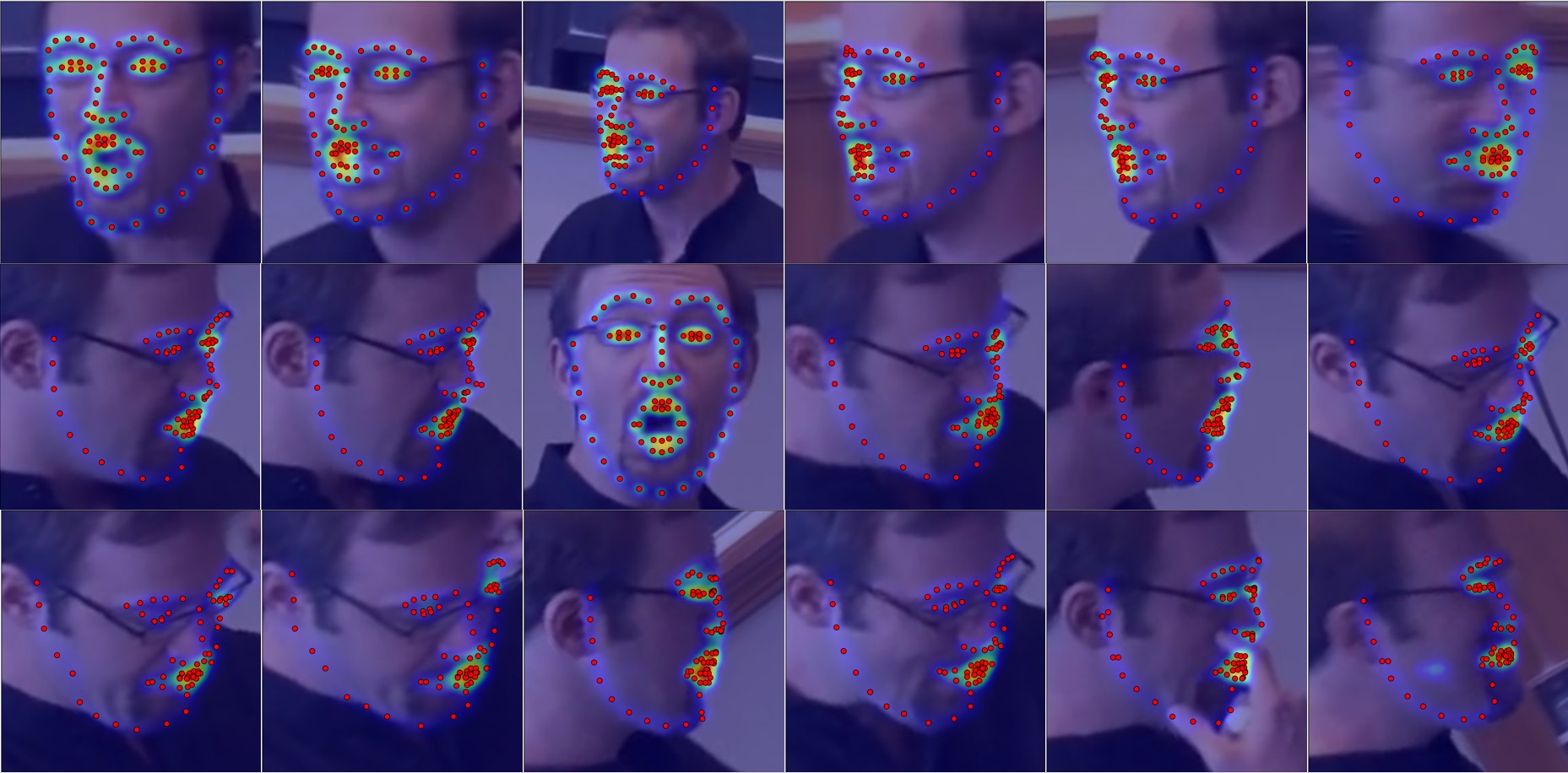}}
  \caption{Response maps generated by MHM on two challenging videos from 300VW Scenario3 (Video ID: 411 and 557). The response maps are invariant to large pose variation and robust under occlusion and fast motion.}\label{pic:300VWdemo}
\end{figure}

\begin{table*}[ht!]
\begin{center}
\begin{tabular}{c|c c|c c|c c}
\hline
&\multicolumn{2}{c}{Scenario1}& \multicolumn{2}{c}{Scenario2} & \multicolumn{2}{c}{Scenario3}\\
\hline
Method & AUC & Failure Rate ($\%$) & AUC & Failure Rate ($\%$) &AUC & Failure Rate ($\%$) \\
\hline\hline
Yang~\etal    &0.791 &2.400 &0.788 &0.322 &0.710 &4.461\\
Xiao~\etal    &0.760 &5.899 &0.782 &3.845 &0.695 &7.379\\
\hline\hline
MDNET + CFSS + Kalman  &0.784 &1.754 &0.783 &0.341 &0.713 &7.466\\
MTCNN + CFSS + Kalman  &0.734 &8.507 &0.725 &8.518 &0.726 &5.685\\
MTCNN + CFSS + previous&0.748 &6.055 &0.760 &2.717 &0.726 &4.388\\
\hline\hline
Our method     &0.847 & 0.290 &0.838 &0.033 &0.769 &0.972\\
Kalman smooth&0.849 & 0.285 &0.842 &0.030 &0.7734&0.889\\
\hline\hline
\end{tabular}
\end{center}
\caption{Landmark localisation results on three categories of the 300VW test sets using 68 landmarks. Accuracy is reported as the Area
Under the Curve (AUC) and the Failure Rate of the Cumulative Error Distribution of the RMS point-to-point error normalised with the diagonal of the ground truth bounding box~\cite{chrysos2016comprehensive}.}\label{table:300VW}
\end{table*}

In Figure~\ref{pic:300VWdemo}, we select some frames from most challenging videos in Scenario3 and show their corresponding response maps for visualisation purpose. The response maps of proposed method is very robust under large pose variation (yaw + pitch angles) and occlusion. In addition, response maps of invisible face parts are also reasonable, which indicates an implicit facial shape constraint within our method. 

\subsection{Face Detection}

\begin{figure}
  \centering
 \subfigure[Evaluation on FDDB]{
    \label{fig:FDDB} 
    \includegraphics[width=0.5\textwidth]{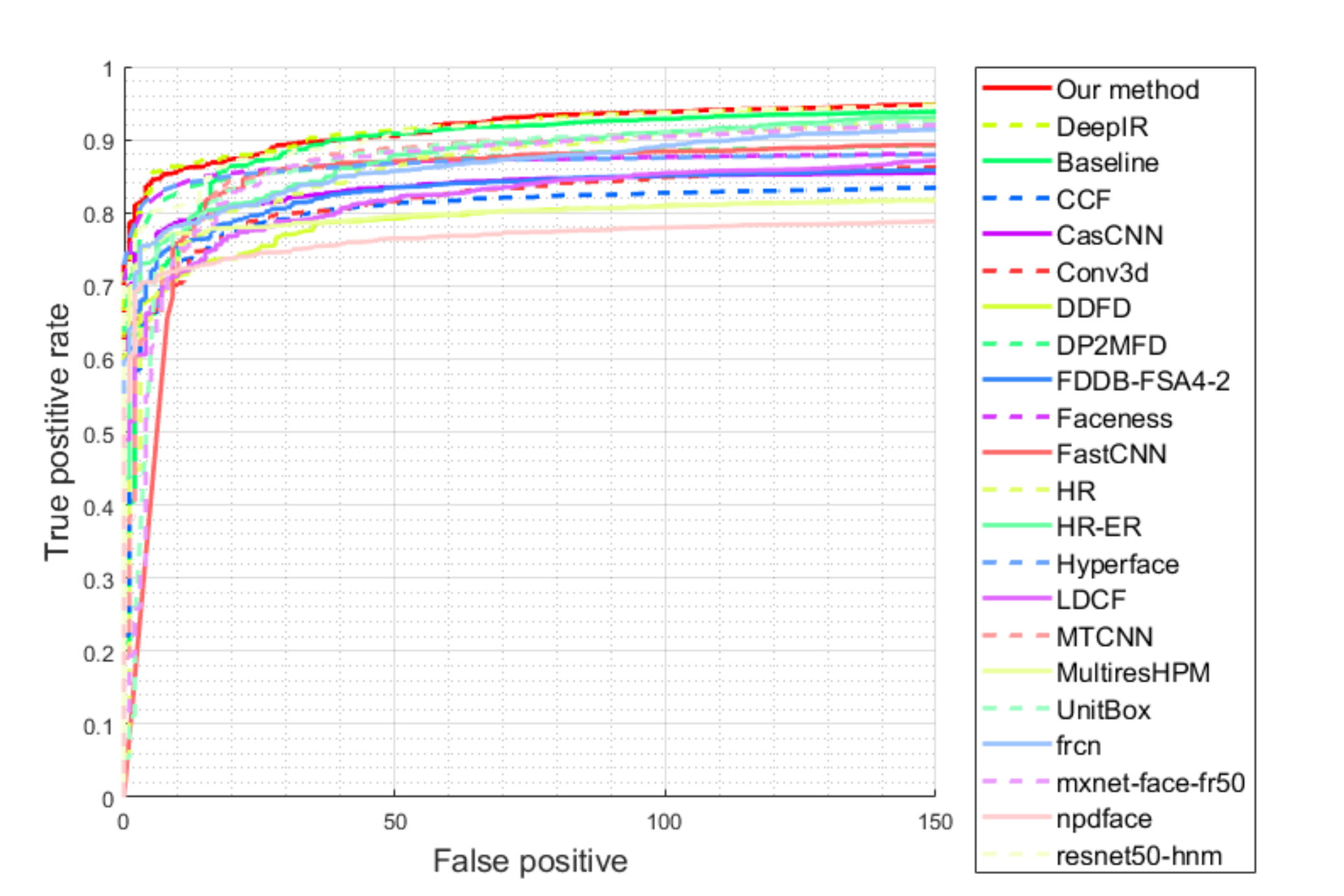}}
  \subfigure[Hard False Positives]{
    \label{pic:FDDB_FP} 
    \includegraphics[width=0.4\textwidth]{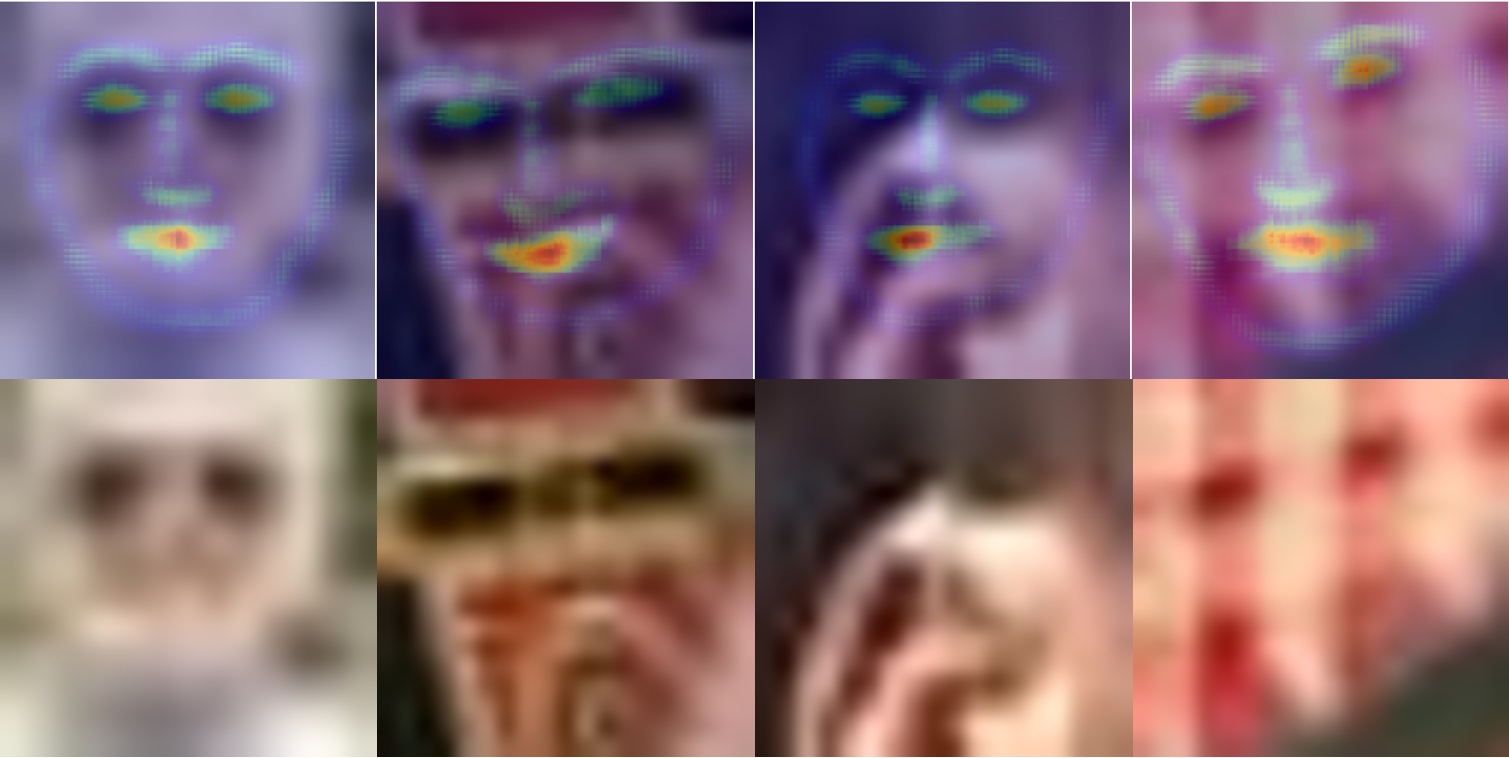}}
 \caption{(a) Face detection results on FDDB. Our method utilises the joint multi-view response maps to remove high score false positives. (b) Some interesting hard false positives from FDDB that even can not remove by our classifier. The resolution of these regions are about $20\times20$, and the structure and contour are very similar to a human face. The first row is covered with the predicted response maps, and the second row is the enlarged image crops.}
\label{fig:FDDB_result}
\end{figure}

We evaluate the effectiveness of the multi-view response maps to remove high score false positives and obtain a state-of-the-art result on the FDDB dataset. As in~\cite{chen2016supervised}, we review the annotation of FDDB~\cite{jain2010fddb}, and add 67 unlabelled faces in FDDB dataset to make sure all the false alarms are correct. We enlarge FDDB images by 1.6, and the average resolution is about $639\times604$. We test the model on a single NVIDIA GTX Titan X (Pascal) GPU setting minimum face as 20.
As shown in Table~\ref{table:FDDB_recall_precision} and Figure~\ref{fig:FDDB},
we observe the improvement of recall within the high precision section (150 false positives, precision rate $97.1\%$).
The baseline method refers to our re-implementation of MTCNN~\cite{zhang2016joint}, due to adopting additional labelled faces from AFLW, our implementation is slightly better than the original MTCNN. Our method {\em th1} sets a higher thresholds (0.6, 0.7, 0.7, 0.7) for cascaded classifiers, while our method {\em th2} employs a lower thresholds (0.5, 0.5, 0.3, 0.7). As can be seen from Table~\ref{table:FDDB_recall_precision} and Table~\ref{table:FDDB_time}, the setting of {\em th2} is slightly better than {\em th1}, but increases the running time from 49.8 ms to 62.9ms per image. The proposed joint multi-view response maps contribute to removing high score false positives from previous cascade classifiers. At the precision rate of $99.9\%$, the proposed method improves the recall from $65.1\%$ to $84.5\%$. At the precision rate of $99\%$, the proposed method improves the recall from $89.9\%$ to $90.5\%$. The result is obviously higher than HR-ER~\cite{hu2016finding} and
Conv3D~\cite{li2016face}, and comparable with the best academic face detectors,
\eg STN~\cite{chen2016supervised}, Xiaomi~\cite{wan2016bootstrapping}, and DeepIR~\cite{sun2017face}. After investigating our false positives, we surprisingly find some tiny regions (shown in Figure~\ref{pic:FDDB_FP}) that can hardly be removed by our method, since they have very similar appearance and structure of the face, and may only be resolved by context-based model.

\begin{table}
\begin{center}
\begin{tabular}{c|c|c|c}
\hline
False Positives      & 5 & 50 & 150  \\
\hline
Precision Rate       & $99.9\%$ & $99\%$ & $97.1\%$\\
\hline\hline
Our method {\em th1} & 84.3 & 90.4 & 90.5 \\
Our method {\em th2} & 84.5 & 90.5 &94.8 \\
Baseline & 65.1 & 89.9 & 92.4 \\
\hline\hline
MTCNN~\cite{zhang2016joint} & 64.2 & 88.8 & 91.8 \\
HR-ER,CVPR17~\cite{hu2016finding} & 73.1 & 87.9 & 93.1\\
Conv3D,ECCV16~\cite{li2016face} & 66.1 & 81.6 & 86.2 \\
STN,ECCV16~\cite{chen2016supervised} & 88.3 & 90.3 & 91.5 \\
Xiaomi~\cite{wan2016bootstrapping} & 78.6 & 90.8 & 94.6 \\
DeepIR~\cite{sun2017face} & 82.7 & 91.2 & 94.7 \\
\hline
\end{tabular}
\end{center}
\caption{Recall rate comparison with the state-of-the-art face detectors on FDDB within the high precision rate section (150 false positives, $97.1\%$).}\label{table:FDDB_recall_precision}
\end{table}

\begin{table}
\begin{center}
\begin{tabular}{c|c|c|c|c}
\hline
    & proposal & CLS2 & CLS3 & {\bf CLS4}   \\
\hline\hline
threshold& 0.6 & 0.7 & 0.7 & 0.7   \\
output boxNum&194.29&15.72&1.84&1.776\\
time(ms)&11.45&2.41&1.74&4.27\\
recall&97.76&95.17&90.97&90.60\\
precision& 0.91 & 11.03 & 89.82 &92.72\\
\hline\hline
threshold&0.5&0.5&0.3&0.7\\
output boxNum&265.27&25.16&2.65&1.784\\
time(ms)&11.89&2.88&2.11&6.15\\
recall&98.30&97.87&95.44&95.10\\
precision & 0.67 &7.07& 65.51&96.89\\
\hline\hline
\end{tabular}
\end{center}
\caption{Output face box number and computation time of each step of our detector under two different threshold setting. Time consumption on image resize and Non-Maximum Suppression (NMS) take about 7.5ms. CLS4 is after multi-view face alignment step (12.21ms per face). For {\em th1}, the mean running time is about 49.8ms per image. For {\em th2}, the mean running time is about 62.9ms per image.}\label{table:FDDB_time}
\end{table}

\begin{figure*}[htp!]
  \centering
  \subfigure[Whole]{
    \label{fig:subfig:Whole} 
    \includegraphics[width=0.32\textwidth]{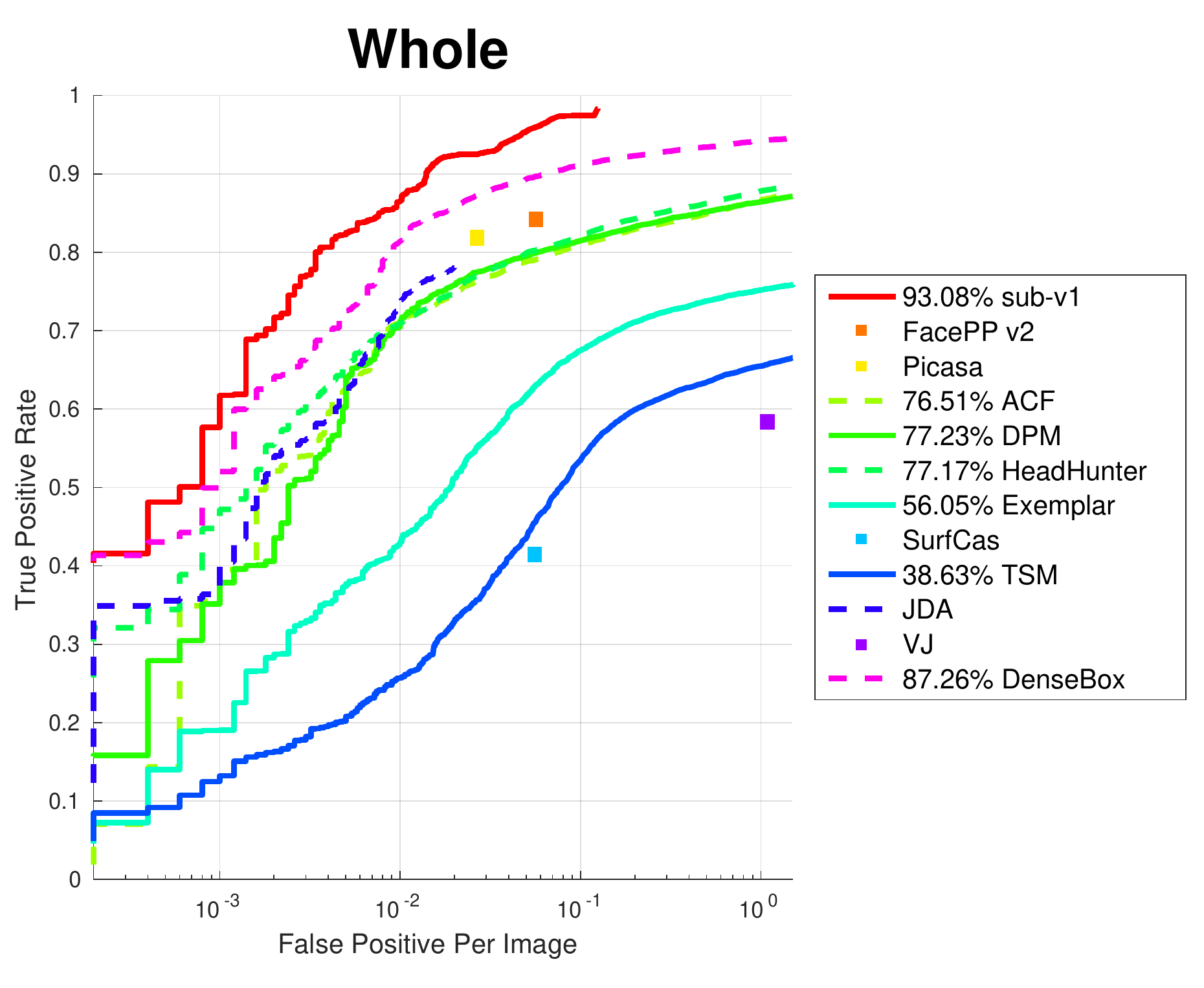}}
 \subfigure[Easy]{
    \label{fig:subfig:Easy} 
    \includegraphics[width=0.32\textwidth]{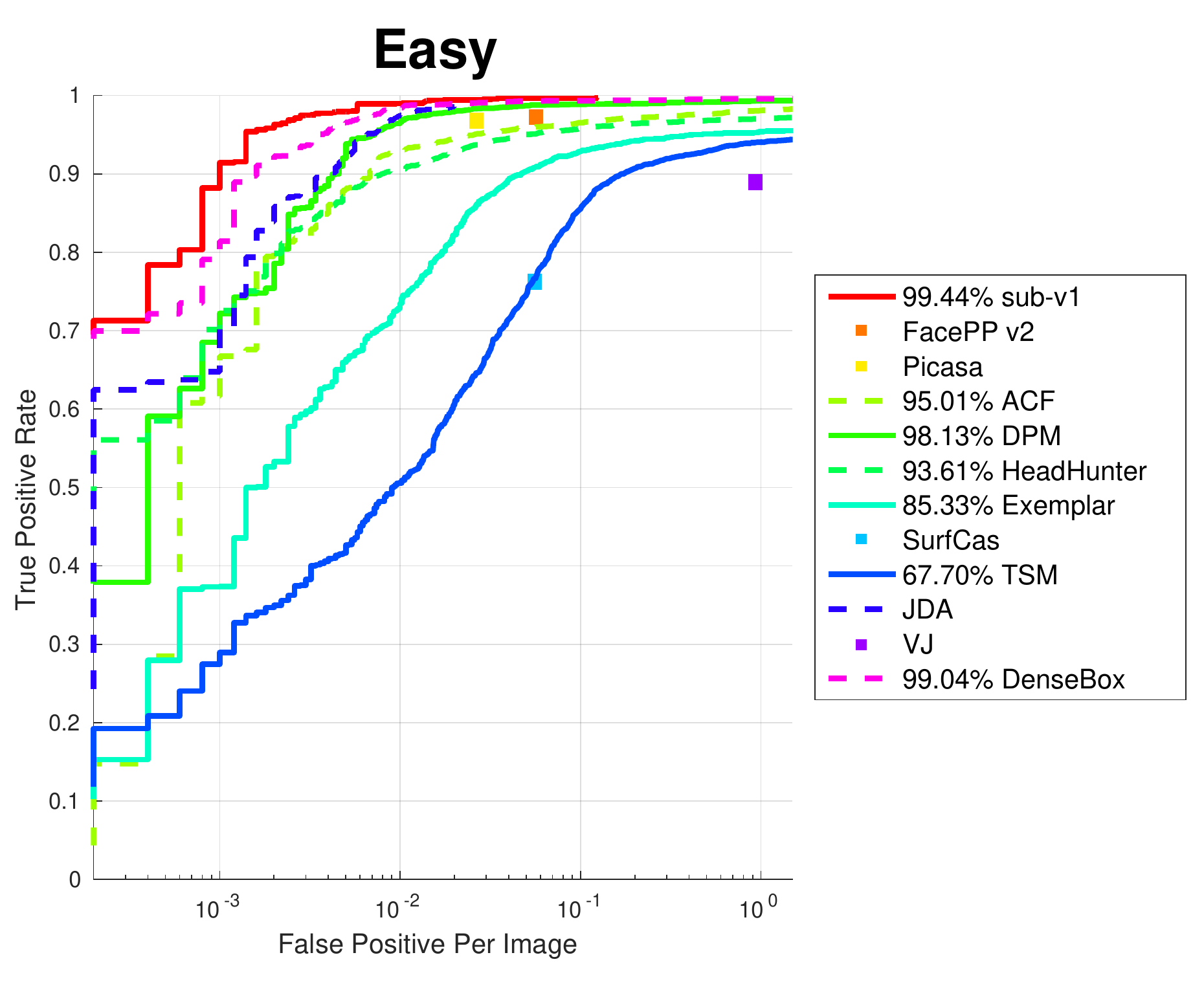}}
  \subfigure[Hard]{
    \label{fig:subfig:Hard} 
    \includegraphics[width=0.32\textwidth]{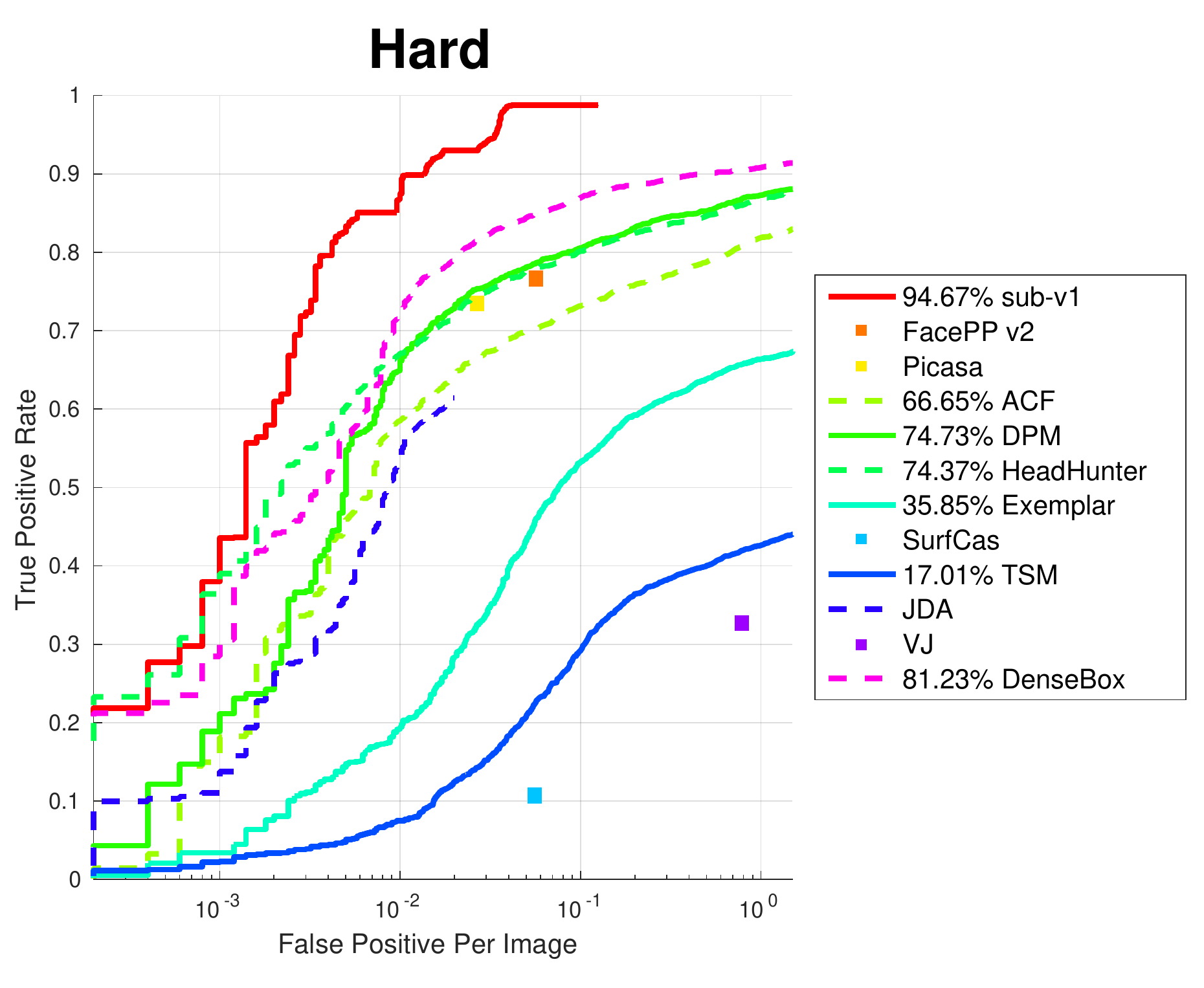}}
 \subfigure[Yaw angle $<$ 20]{
    \label{fig:subfig:Low resolution} 
    \includegraphics[width=0.32\textwidth]{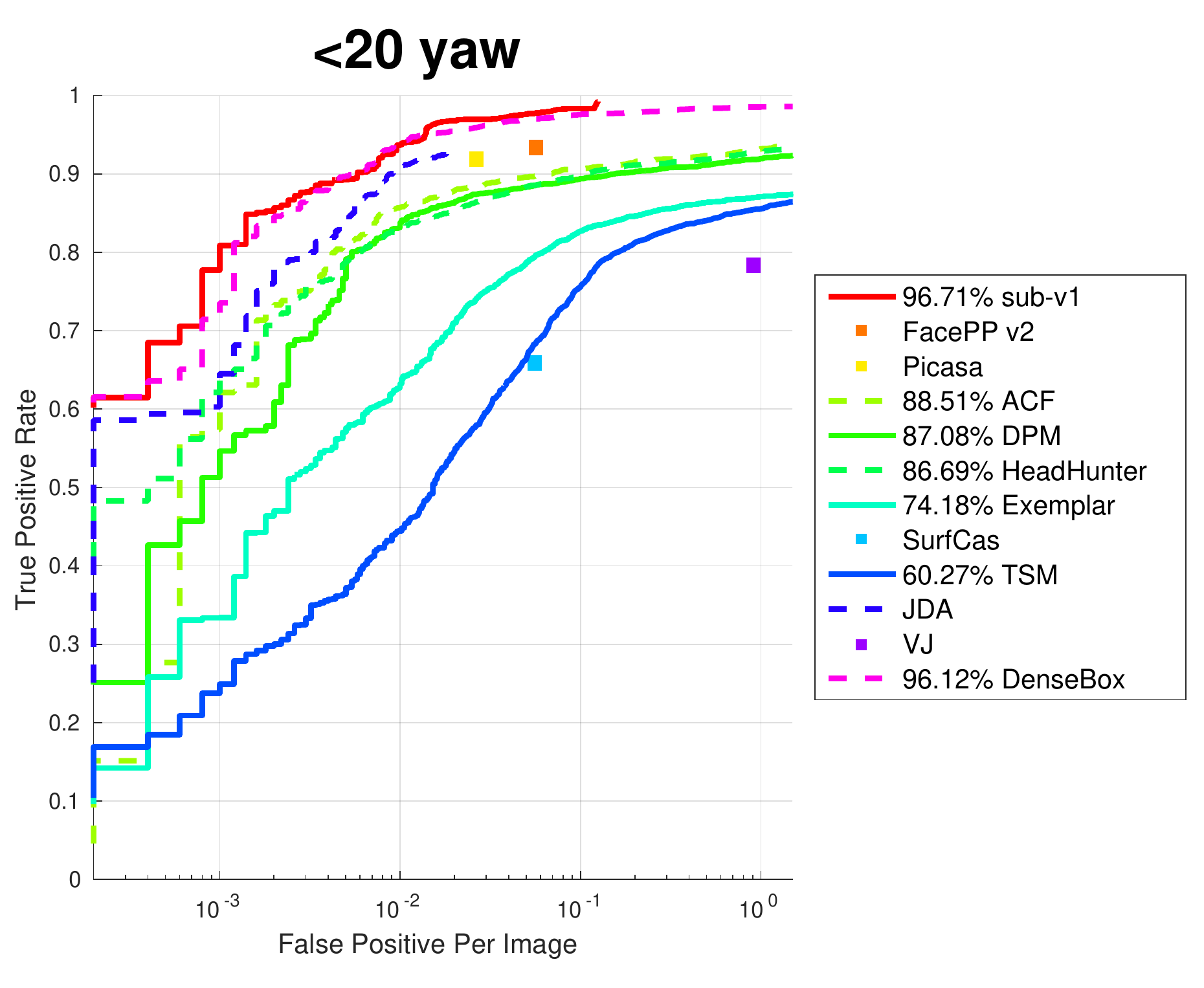}}
  \subfigure[Yaw angle $\in$ (20,40)]{
   \label{fig:subfig:High resolution} 
    \includegraphics[width=0.32\textwidth]{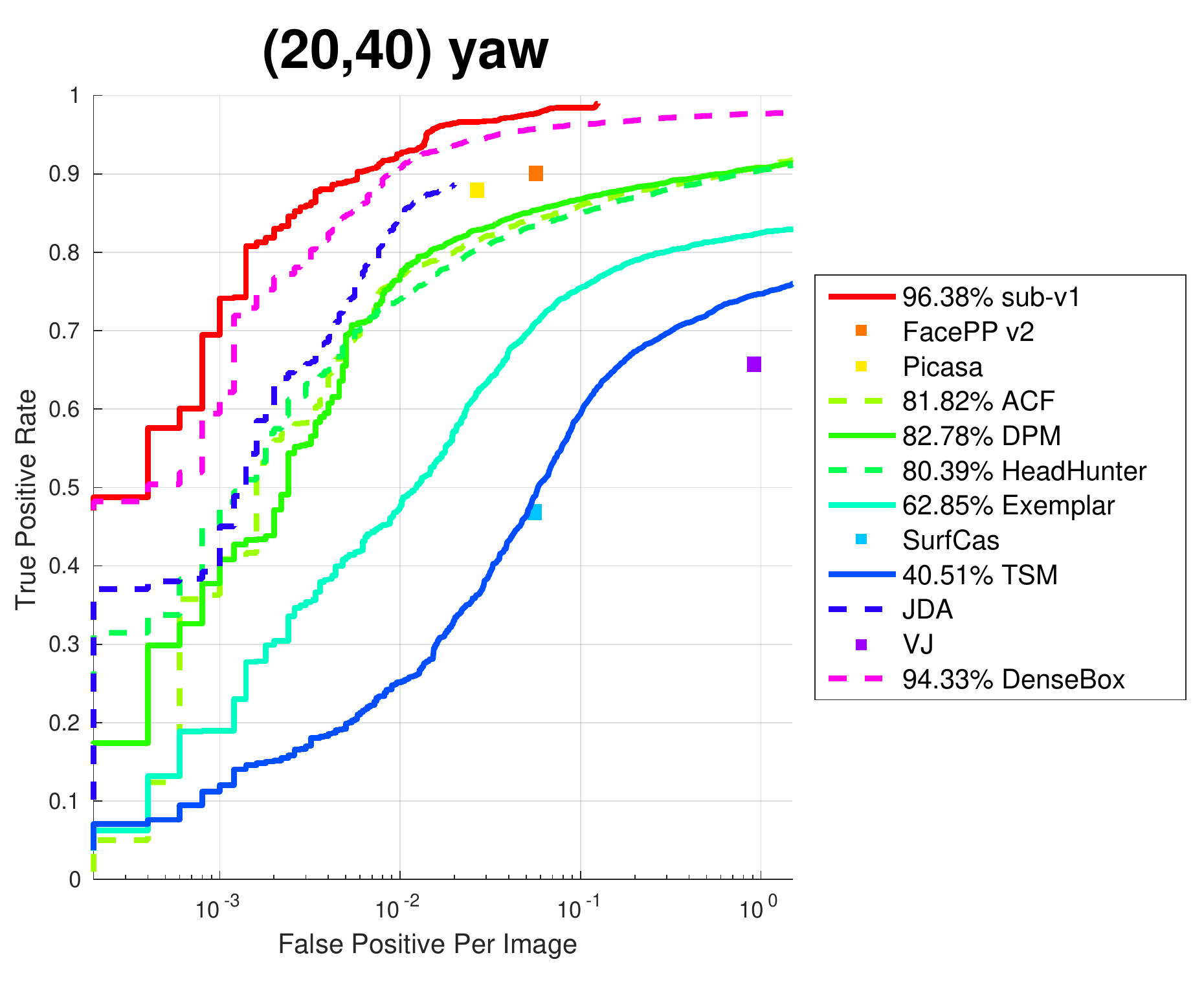}}
  \subfigure[Yaw angle $>$ 40]{
    \label{fig:subfig:Pose-yaw} 
    \includegraphics[width=0.32\textwidth]{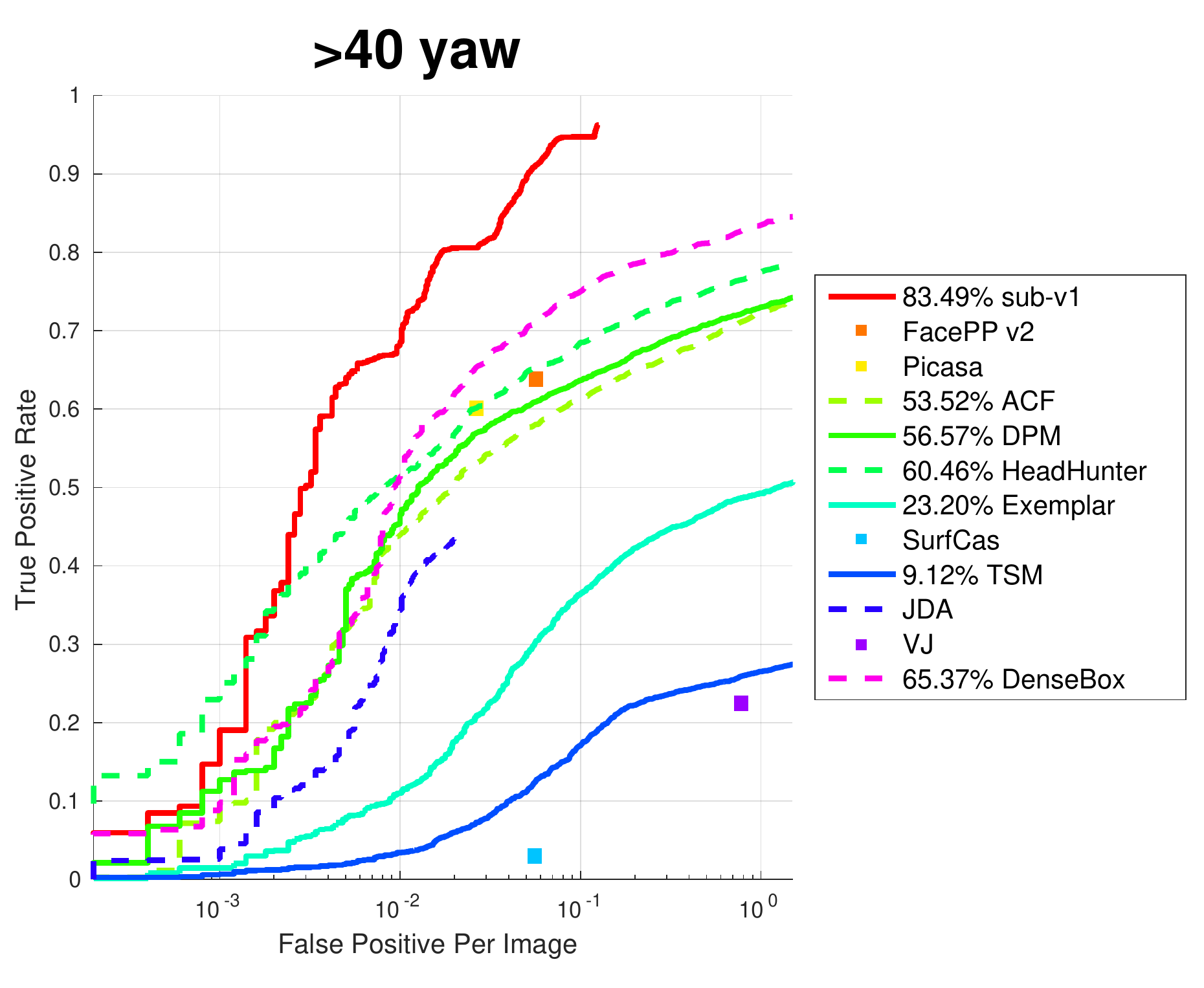}}
 \caption{Fine-grained evaluation on MALF dataset.}
\label{fig:MALF} 
\end{figure*}

We also submitted our face detection results to {http://www.cbsr.ia.ac.cn/faceevaluation/} and obtained the true positive vs. false positive curve on MALF. In Figure~\ref{fig:MALF}, our submission is named ``$sub\_v1$'' and the threshold setting is (0.5,0.5,0.3,0.7). We compared with the off-the-shelf face detectors including HeadHunter~\cite{mathias2014face}, ACF~\cite{yang2014aggregate}, DPM~\cite{mathias2014face}, JDA~\cite{chen2014joint}, and DenseBox~\cite{huang2015densebox}. The proposed method obtains the best performance on MALF compared to the best academic algorithms including cascade models (HeadHunter~\cite{mathias2014face}, ACF~\cite{yang2014aggregate}, JDA~\cite{chen2014joint}), structure models (DPM, JDA) and the structure-constrained deep model (Densebox). We also outperform the big data driven commercial models such as the FacePP-v2 and Picasa algorithms. Compared to the state-of-the-art method DenseBox, our joint multi-view response maps achieve a significantly better detection result in large pose data (yaw angle $>$ 40 degrees). A similar improvement could also be observed on the ``hard'' subsets.

\section{Conclusion}
In this paper, we proposed a coarse-to-fine multi-view face alignment method where a face detector is used to estimate a coarse estimate of the facial shape using a small subset of landmarks and then after removing similarity transformations a refining subsequent step is performed that estimates the high-resolution facial shape of each person. We formulate a novel multi-view hourglass model which tries to jointly estimate both semi-frontal and profile facial landmarks, and the joint training model is stable and robust under continuous view variations.
We demonstrate huge improvement over the state-of-the-art results in the latest benchmarks for face alignment such as 300W, COFW and the latest Menpo Benchmark. We also demonstrate state-of-the-art results for the deformable face tracking on the 300VW benchmark and face detection on FDDB and MALF datasets.

\ifCLASSOPTIONcaptionsoff
  \newpage
\fi

\bibliography{egbib}
\bibliographystyle{IEEEbib}

\end{document}